\documentclass{article} % For LaTeX2e
\usepackage{iclr2026_conference,times}

\usepackage{amsmath,amsfonts,bm}

\def\eqref#1{equation~\ref{#1}}
\def\1{\bm{1}}

\DeclareMathAlphabet{\mathsfit}{\encodingdefault}{\sfdefault}{m}{sl}
\SetMathAlphabet{\mathsfit}{bold}{\encodingdefault}{\sfdefault}{bx}{n}

\PassOptionsToPackage{hyphens}{url}\usepackage{hyperref}
\usepackage{url}
\usepackage[utf8]{inputenc} % allow utf-8 input
\usepackage[T1]{fontenc}    % use 8-bit T1 fonts
\usepackage{url}            % simple URL typesetting
\usepackage{booktabs}       % professional-quality tables
\usepackage{amsfonts}       % blackboard math symbols
\usepackage{nicefrac}       % compact symbols for 1/2, etc.
\usepackage{microtype}      % microtypography
\usepackage{xspace}
\usepackage{graphicx}
\usepackage[many]{tcolorbox}
\usepackage{setspace}       % for LINE SPACING
\usepackage{multicol}       % for MULTICOLUMNS
\usepackage{minted}
\usepackage{float}

\newcommand{\sys}{{\scshape safe}\xspace}

\definecolor{main}{HTML}{0d6efd}    % setting main color to be used
\newcounter{takeawaynum}
\newtcolorbox{takeaway}{
    title = Takeaway \thetakeawaynum,
    before title app = \stepcounter{takeawaynum}\vspace{2pt},
    fonttitle = \bfseries,
    boxrule = 1.5pt,
    colframe = main,
    rounded corners,
    arc = 5pt   % corners roundness
}

\title{SAFE: A Novel Approach to AI Weather Evaluation through Stratified Assessments of Forecasts over Earth}

\author{%
  Nick Masi\\
  Department of Computer Science\\
  Brown University\\
  \texttt{nicholas\textunderscore  masi@alumni.brown.edu} \\
  \And
  Randall Balestriero \\
  Department of Computer Science \\
  Brown University \\
  \texttt{randall\textunderscore balestriero@brown.edu} \\
}

\iclrfinalcopy % Uncomment for camera-ready version, but NOT for submission.
\begin{document}

\maketitle

\begin{abstract}
The dominant paradigm in machine learning is to assess model performance based on average loss across all samples in some test set. This amounts to averaging performance geospatially across the Earth in weather and climate settings, failing to account for the non-uniform distribution of human development and geography. We introduce Stratified Assessments of Forecasts over Earth (\sys), a package for elucidating the stratified performance of a set of predictions made over Earth. \sys integrates various data domains to stratify by different attributes associated with geospatial gridpoints: territory (usually country), global subregion, income, and landcover (land or water). This allows us to examine the performance of models for each individual stratum of the different attributes (e.g., the accuracy in every individual country). To demonstrate its importance, we utilize \sys to benchmark a zoo of state-of-the-art AI-based weather prediction models, finding that they all exhibit disparities in forecasting skill across every attribute. We use this to seed a benchmark of model forecast fairness through stratification at different lead times for various climatic variables. By moving beyond globally-averaged metrics, we for the first time ask: where do models perform best or worst, and which models are most fair? To support further work in this direction, the \sys package is open source and available at \url{https://github.com/N-Masi/safe}.

% Anonymous link: \url{ https://anonymous.4open.science/r/safe-E7C7}.

\end{abstract}

\section{Introduction}

Artificial intelligence weather prediction (AIWP) models, alternatively machine learning weather prediction (MLWP) models or neural weather models (NWM), are becoming increasingly competitive with traditional numerical weather prediction (NWP) models. All of these approaches are typically used in making medium-range weather forecasts (interchangeably, ``prediction''). The range of a forecast is determined by its lead time $\tau$. When a weather prediction model is fed the state of variables at time $d$, its task is to predict the state of those variables (or some subset of them) at time $d+\tau$. There is no consistent definition for medium-range, with the European Centre for Medium-Range Weather Forecasts (ECMWF) defining it as any prediction made with $\tau$ (or $n\times\tau$ if taking an autoregressive rollout of $n$ steps) within 0–15 days \citep{ecmwfdef}, while other sources more narrowly define it as 3–7 days \citep{amsdef, nwsdef}. AIWP models are seeing increasing adoption in interfaces where they provide these medium-range forecasts, from Google's Weather app \citep{leffer24sciam} to various experimental models at the National Oceanic and Atmospheric Administration (NOAA) \citep{noaa2025wofscast, noaa2025graphcast}.

Root mean square error (RMSE) is the preeminent metric used in assessing the quality of AIWP models \citep{radford2025comparison, rasp2020weatherbench}. The general form of RMSE is shown in \autoref{rmse}, where $Y$ is the set of all ground truth variable values that a model is trying to predict, and $\hat{y}$ is the model's prediction for each corresponding $y\in Y$. Every $y$ is the value of some variable (e.g., temperature) at some point in time $d\in D$, longitude $i\in I$, latitude $j\in J$, and, for certain atmospheric variables, vertical level $v\in V$. 

\begin{equation} \label{rmse}
    \sqrt{\frac{\Sigma_{y\in Y}(\hat{y}-y)^2}{|Y|}}
\end{equation}

There are various approaches for how different models handle being able to make predictions at different lead times. The naive approach is to train a model with the ability to predict some fixed $\tau'\in T$ amount of time in the future, where $T$ is a set of durations. This allows the model to forecast the weather with temporal resolution of $\tau'$ (i.e., multiples of $\tau'$ after the timestamp of the input variables) through autoregressive rollout. This is the approach taken by Keisler \citep{keisler2022forecasting}, the Spherical CNN \citep{esteves23sphericalcnn}, FourCastNet \citep{pathak2022fourcastnet}, and the spherical Fourier neural operator (SFNO) \citep{bonev2023sfno}, all with $\tau'= 6$ hours. Pangu-Weather \citep{bi2022pangu} trains four different models, each with a different, fixed lead time. This is used in tandem with a greedy algorithm that minimizes the number of autoregressive steps that need to be taken to make a prediction at any given lead time (which must be a multiple of their smallest lead time model). FuXi \citep{chen2023fuxi} uses a cascaded set of three different models that cover different ranges of lead times.
 
The square of RMSE, mean squared error (MSE), frequently referred to as the $L2$ loss, is often used as a training objective. This is the case for Spherical CNN \citep{esteves23sphericalcnn} and GenNet \citep{lopez2023gennet}. GraphCast \citep{lam2023learninggraphcast} and GenCast \citep{price2023gencast} use weighted MSE loss functions. Keisler takes a weighted sum of MSE values \citep{keisler2022forecasting}. NeuralGCM \citep{kochkov2024neuralgcm} has a five-term loss function, each of which is a variation of MSE. FuXi \citep{chen2023fuxi} uses the mean absolute error (MAE, the $L1$ counterpart of MSE).

The underlying commonality across all of these functions is that they completely reduce across the spatial dimensions $I$ and $J$. One issue with spatial averaging as the loss function is the resulting ``double penalty" that arises when predictions for high resolution events are even slightly spatially displaced, incurring the penalization for both that faulty prediction and the lack of prediction at the true location \citep{gilleland_intercomparison_2009}. This encourages models to blur their predictions, dropping these highly localized events \citep{lam2023learninggraphcast}. However, neglecting to predict these outlier events can have dramatic real-world consequences. For example, improved accuracy of extreme heat predictions has been found to reduce mortality \citep{shrader_fatal_2023}. Another issue with spatial averaging for evaluation is that it becomes unknown precisely where models are and are not performing well. Accordingly, it is impossible to know whether they can be trusted at inference time in a given location. With \sys, we aim to uncover spatial disparities in performance by separating the spatial dimensions into different strata and calculating performance within each.

\begin{takeaway}
The state of the art of AI weather prediction relies on spatially-averaged objective functions and evaluation metrics. These de-emphasize high-frequency events despite the fatal consequences of losing this predictive power. They also mask disparities that exist in where models perform well.
\end{takeaway}

\section{Related work} \label{related-work}

WeatherBench 2 (WB2) \citep{rasp2024weatherbench2} is an existing benchmark that assesses the spatially-averaged error of models using weather data from ERA5, ECMWF's most modern reanalysis dataset \citep{hersbach2020era5}. It provides functionality to get per-region RMSE, but these regions are coarse-grained and exclusively rectangular, making them unusable for the real-world attributes we care about.

Stable equitable error in probability space (SEEPS) \citep{rodwell2010seeps} is a metric that was introduced to assess the quality of precipitation forecasts in particular. In the original paper \citep{rodwell2010seeps}, the authors perform region-specific analysis of forecasts in South America, Europe, and the extropics. Again, however, the region shapes are defined with crude, rectangular boundaries ([70$^\circ$W–35$^\circ$W, 40$^\circ$S–10$^\circ$N], [12.5$^\circ$W–42.5$^\circ$E, 35$^\circ$N–75$^\circ$N], and [above 30$^\circ$N or below 30$^\circ$S], respectively).

NeuralGCM also calculated per-region RMSE for T850 and Z500 \citep[Supp. Mat. Fig. S14–S16]{lam2023learninggraphcast}, borrowing region definitions from ECMWF scorecards. There are 20 of these regions, 3 that are hemispheric (North, Tropical, and Southern) and 17 geographic. These regions are overlapping and include oceans, but the geographic regions miss considerable sections of populated landmass (including but not limited to significant portions of Central America, Eastern Africa, Brazil, California, and the island of New Guinea). The hemispheric regions cover the whole globe, with the Tropical region bounded by the $\pm20^\circ$ latitude lines.

In contrast, the regions used within \sys cover all landmass (including islands) across the Earth and are carefully crafted to not include oceanic landcover. This more aptly captures metrics for where fairness in weather forecasts matters most: the places where people live. Our regions are non-overlapping, except at their borders where gridpoint polygons stretch over the border (this being an artifact of finite resolution).

\begin{takeaway}
    Existing approaches to stratify AIWP model performance are rare and at best utilize crude rectangular boundaries, operating only on the subregion attribute.
\end{takeaway}

\section{SAFE}

In this paper we create a framework for performing Stratified Assessments of Forecasts over Earth (\sys). This tool enables  stratification by various geographically-related attributes, allowing the user to see the fine-grained quality of a set of predictions when broken down by the different constituent groups, or strata, of each attribute. We leverage \sys to benchmark the fairness of existing AIWP models. Despite the life or death impacts of weather forecasts and concrete evidence that existing forecasts provided by the National Weather Service have error rates that vary across the geography of the United States \citep{wapo_2024}, there is little existing work that investigates model error spatially (see: \autoref{related-work}).

\begin{takeaway}
    We introduce \sys, an open source python library that integrates different data sources and facilitates stratified fairness evaluations of AI weather and climate models.
\end{takeaway}

\subsection{Data sources} \label{data-sources}

Within \sys, we provide the ability to investigate different attributes: territory, global subregion, income, and landcover. The strata within the territory attribute is typically the country which a gridpoint is located within, though there are some sub-national or not universally recognized territories. Territory borders are pulled from the geoBoundaries Global Administrative Database \citep{runfola2020geoboundaries}. Any gridpoint overlapping with any land will be classified as ``land'' for the landcover attribute and otherwise as ``water''. Global subregions follow the United Nation's classifications over territories \citep{un99seriesm49}. The income stratum of a gridpoint is one of ``high income'', ``upper-middle income'', ``lower-middle income'', or ``low-income'' as defined by the World Bank's classification for the gridpoint's encompassing territory \citep{worldbankincome}; the World Bank uses the gross national income (GNI) per capita of the territory, calculated using the Atlas methodology. The polygons associated with each strata are accessed through the MIT-licensed pygeoboundaries\textunderscore geolab package \footnote{\url{https://github.com/ibhalin/pygeoboundaries}}. This package is a python wrapper for the geoBoundaries Global Administrative Database \citep{runfola2020geoboundaries}, which itself is made available under a open license CC-BY 4.0. 

\subsection{Methods}
 
\subsubsection{Stratification} \label{stratum-membership-methods}

Forecasts made over the Earth are associated with specific (longitude, latitude) coordinates, or ``gridpoints'' on the Earth. Each pair of coordinates is converted into the polygon that is centered on the gridpoint but which covers all the quadrilateral surface area defined by extending its borders to the midpoint with its neighbors in both the longitude and latitude directions. To unify the coordinate system across all integrated data sources, latitude ranges [-90, 90] with index 0 at -90, and longitude [-180, 180] but with index 0 at 0 and a wraparound from 180 to -180 in the middle. This is because polygons and associated attribute metadata sourced from pygeoboundaries\textunderscore geolab follows this coordinate system, and it is easier to bring the other tabular data into conformance than modify this.

The forecasts for a gridpoint's polygon are associated with all of the strata that have any polygon which intersects it. While this will double count some gridpoints towards different strata, measures are taken so that no single gridpoint counts more than once within a given strata. The double counting that does occur is in line with the philosophy of \sys, as the alternative is that—without high enough resolution—there will be strata for which no data is recorded, rendering them invisible and left out of fairness assessments. Importantly, this ``double counting'' is a different phenomenon from the ``double penalty'' described by \citet{gilleland_intercomparison_2009}. In total, there are 231 territory, 23 subregion, 4 income, and 2 landcover strata. Of the 231 territories, 213 have an associated income strata. 76 are classified as high-income, 57 as upper-middle-income, 45 as lower-middle-income, and 34 as low-income. Subregions vary from having 1 territory (Antarctica) to 25 (Caribbean). More details on the strata are in \autoref{attribute-strata-details}.

\subsubsection{Area weighting}

In calculating the loss function for training it is common to weight the (squared if $L2$) difference in variable prediction and ground truth by the area of the gridpoint cell the forecast was made at before averaging. This weight varies with latitude. The reason for latitude weighting is that, when using an equiangular gridding, the gridpoints are closer together near the poles than they are at the equator. This results in a higher density of samples per area at the poles, which left unaccounted for could cause the model to overfit to forecasting polar weather. 

Complicating the matter, Earth is an oblate spheroid with an equatorial radius of 6,378,137m and a smaller polar radius of 6,356,752m. However, no existent python library known to the authors takes this into account to get the precise surface area of equiangular grid cells on Earth's surface. The assumption of a spherical Earth yields surface areas near the poles that are still greater than they are in reality, meaning the very problem latitude weighting aims to address persists. The standard solution would be to convert the cells to vector data and get the area of polygons. However, virtually every approach, both training \citep{lam2023learninggraphcast, keisler2022forecasting, bi2022pangu, kochkov2024neuralgcm, pathak2022fourcastnet, bonev2023sfno} and benchmarking \citep{rasp2024weatherbench2, Leeuwenburg_scores_A_Python_2024}, make the simplifying assumption of a perfectly spherical Earth. WB2 takes this approach in computing its metrics as well \citep{rasp2024weatherbench2}. As part \sys, we have provided a utility that can get the surface area of grid cells on the Earth while taking into account its oblate geometry. We use the equation for getting the surface area of oblate spheroid caps from \citet[Eq. 49]{calvimontes2018measurement} which builds on the model developed by \citet{whyman2009oblate}. For testing, the total surface area of the Earth was found with the equation for oblate spheroid surface area from \citet[p. 131]{beyer2018handbook}, yielding an approximation of $\textup{510,065,604,944,206.145m}^2$. A spherical model overestimates the latitude weight (normalized by mean grid cell area) of the polar grid cells (i.e., the most northern or southern grid cells) by 0.7\% with $1.5^\circ$ resolution and by 504\% with $0.25^\circ$ resolution.

In calculating the RMSE as reported throughout this paper, we use these exact surface areas and get the weights by normalizing the grid cell areas by the mean cell area. This same normalization is used in WB2 \citep{rasp2024weatherbench2} and is common in training \citep{pathak2022fourcastnet, bonev2023sfno}.

\begin{takeaway}
    \sys introduces a new state-of-the-art level of accuracy in latitude weighting, a normalization technique used in virtually all AI weather or climate work.
\end{takeaway}

\subsubsection{Metrics} \label{safe-metrics}

\textbf{Model performance metrics.} The main metric utilized in \sys is the latitude-weighted RMSE, which is averaged temporally by initialization time (the timestamp of the climate variables fed into the model) not lead time (the amount of time into the future for which to forecast the state of climate variables at), and averaged spatially within each strata. Unless otherwise specified, reported RMSE refers to this. The anomaly correlation coefficient (ACC) is another evaluation metric that is often used for cross-model comparison. It is the only scale-free metric that is commonly used for this purpose. Like RMSE, ACC is spatially averaged \citep{ecmwf_acc} and would thus benefit from stratified assessment. The fact that the most popular metrics employ spatial averaging underscores the need for \sys. We emphasize RMSE in this work under the same rationale as taken by WeatherBench: the similarity between RMSE and the models' training objectives \citep{rasp2020weatherbench}. Furthermore, RMSE is the predominant metric reported in the literature. In this work we focus on benchmarking deterministic models. Probabilistic, or ensemble, AIWP models have other metrics that can be used such as the continuous ranked probability score (CRPS), but also are commonly evaluated on the RMSE of the ensemble's average prediction.

\begin{figure}
  \centering
  \includegraphics[width=.65\linewidth]{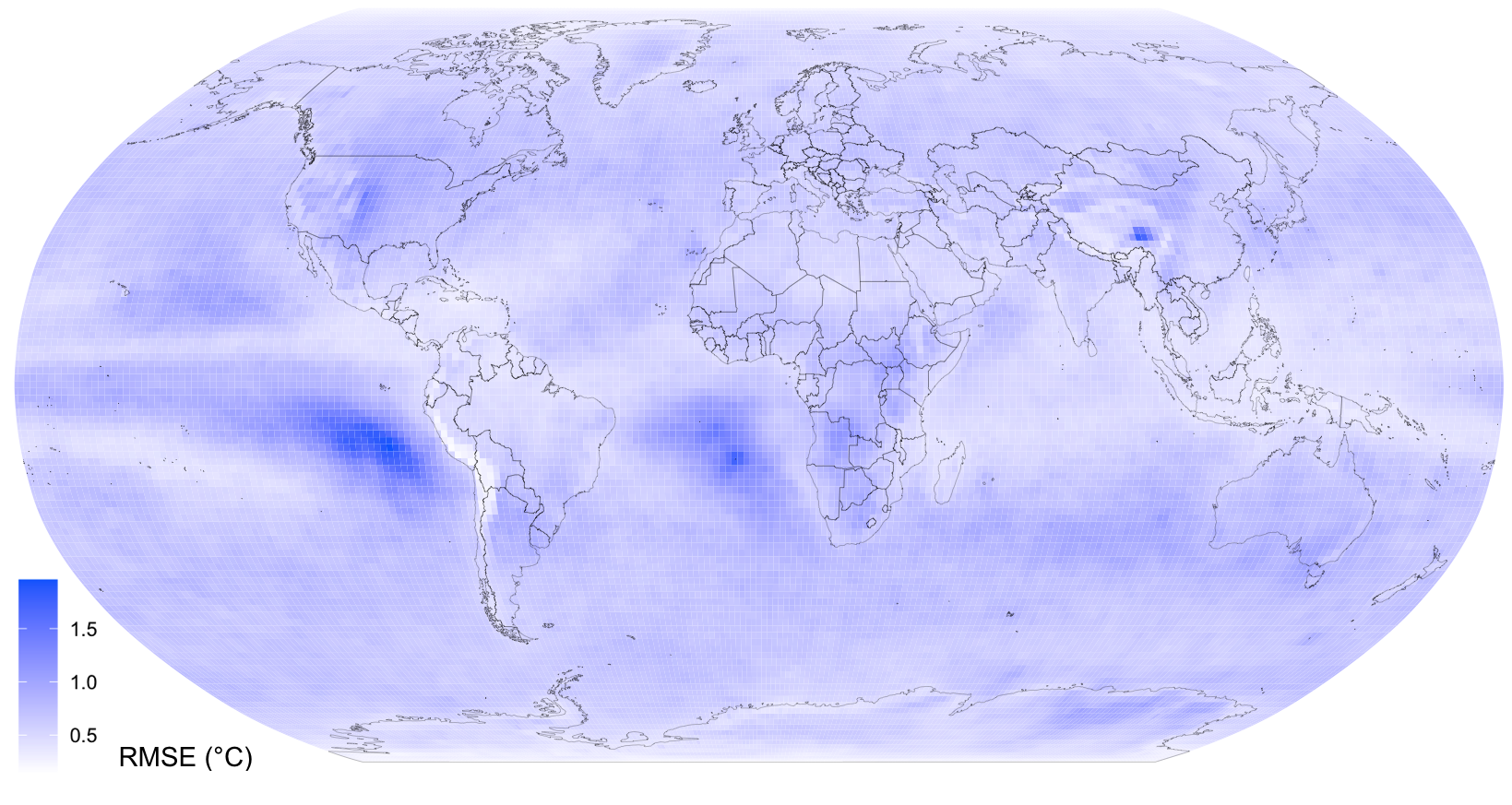}
  \caption{GraphCast displays non-uniform error in temperature prediction. The temporally-averaged gridpoint specific RMSE of temperature predictions at 850hPa (T850) made by GraphCast for every 12 hours in 2020 are shown. Predictions made with 3 day lead time, meaning they predict the temperature 72 hours after the input conditions. Lower RMSE is better. GraphCast inference predictions from WeatherBench 2, ground truth temperature values from ECMWF ERA5. Spatial resolution is 1.5 degrees.}
  \label{graphcast-banner-img}
\end{figure}

We motivate the work of stratified fairness through the spatial disparities that exist in AIWP performance that is visible even on the individual gridpoint level. \autoref{graphcast-banner-img} demonstrates an example of this, showing the unequal performance of GraphCast across the globe at forecasting temperature with $\tau=72$h. The data visualized in this example can be easily accessed with \sys through a call to the \mintinline{python}{safe_earth.metrics.errors.stratified_rmse} function.

\textbf{Fairness metrics.} We define two new metrics for measuring fairness. Both operate on the level of data for individual variables and individual attributes. To start, the RMSE for a model's performance on the given variable is calculated for each strata within the attribute. To characterize the worst-case disparity of each model, we measure (1) the greatest absolute difference in the per-strata RMSEs. To assess the overall nature of the model, we also measure (2) the variance in per-strata RMSEs. An optimally ``fair'' model will have a value of 0 for both metrics, as this would mean it is performing no worse on any strata than any other. These metrics are computed through calls to \mintinline{python}{safe_earth.metrics.fairness.measure_fairness} within \sys.

\sys is easily extensible to incorporate future fairness metrics as they are developed by the theoretical fairness field. Presently, the overwhelming focus of the machine learning fairness community is metrics that apply to binary outcomes, rather than the continuous value we are tracking, and typically in binary (two strata) settings \citep{jui2024fairness, mehrabi2021survey}. This means there is no standard approach for us to take in quantifying fairness as a measure of a continuous outcome that differs across multiple strata per attribute. However, our greatest absolute difference and variance measurements are firmly grounded in the literature that does exist. Many fairness metrics used in both the literature and high context legal settings are also simple differences in performance calculated with subtraction (e.g., statistical parity difference, equal opportunity difference), setting the precedence for our metric.

\begin{takeaway}
    We for the first time introduce \textbf{fine-grain stratification} in the literature. Current approaches use globally-averaged training objectives and evaluation metrics. To start, \sys offers stratification on the attributes of territorial affiliations (country), global subregion, income, and landcover (land or water). \\
    
    We also introduce \textbf{brand new fairness metrics} that are grounded in the existing machine learning fairness field. This empowers many new lines of investigation, such as comparing different models' performance in specific countries or benchmarking model bias.
\end{takeaway}

\section{Benchmarking AIWP forecast fairness: demonstrating SAFE} \label{experiments}

To minimize computational costs, we investigate models with already available predictions. This eliminates the need for model training or inference, reducing the carbon footprint of our research. WB2 provides easily-accessible cloud datasets of ERA5 data and inference runs in the year 2020 for a number of models. Because of the unified access endpoints and resolution, we use the models available through these datasets to begin our investigation. Furthermore, these models are among the most state of the art (by standard metrics such as RMSE and ACC) \citep{raspsotavstimefig}, so it is in fact preferable to study these than retrain our own, potentially inferior models.

\subsection{Forecasts assessed} \label{experimental-details}

In this work with utilize WB2's $1.5^\circ$ resolution equiangular predictions on ERA5. We choose the $1.5^\circ$ resolution ($240\times121$ in terms of longitude by latitude) because it has the most amount of models with provided forecasts at a single common resolution. The forecasts provided are made on ERA5 data from 2020. WB2 retrieved this subset of ERA5 data from ECMWF via the Copernicus Climate Data Store, which makes its products available through an open license.\footnote{\url{https://apps.ecmwf.int/datasets/licences/copernicus/}} WB2 itself is available through an Apache License 2.0.

% Higher resolution forecasts would permit more fine-grained stratification and remediate the double-counting issue discussed in \autoref{stratum-membership-methods}. Indeed, WB2 provides higher resolution than this for some of the models in its zoo. However, benchmarking models against one another is only meaningful when performed at the same resolution. Without this, predictions made at higher resolutions may not get assigned to the same strata.

The models evaluated are GraphCast \citep{lam2023learninggraphcast}, Keisler \citep{keisler2022forecasting}, Pangu-Weather \citep{bi2022pangu}, Spherical CNN \citep{esteves23sphericalcnn}, FuXi \citep{chen2023fuxi}, and NeuralGCM \citep{kochkov2024neuralgcm}; more details on these models are available in \autoref{model-details}. All of the assessed models were trained on ERA5 data, making it an appropriate common benchmark, and none of them included 2020 in their training set. The set of lead times $\tau$ that is common to the provided predictions for all models is every 12 hours up to 10 days, so we assess all models at each of these.

\subsection{Variables} \label{safe-variables}

In line with WeatherBench \citep{rasp2020weatherbench, rasp2024weatherbench2}, we choose as our variables $y$ the atmospheric temperature at 850hPa (``T850'', unit: K) and geopotential at 500hPa (``Z500'', unit: $\textup{m}^2\textup{s}^{-2}$) as benchmark variables for comparing cross-model performance in this experiment. Geopotential is the strength of Earth's gravitational field, so predicting the geopotential at a fixed atmospheric pressure level (500hPa) amounts to predicting the vertical synoptic-scale distribution of pressure in Earth's atmosphere. This knowledge is highly useful in meteorological predictions \citep{lam2023learninggraphcast}. 
% T850 is chosen because the variable of temperature and pressure level closer to the surface make it more tangibly impactful \citep{rasp2020weatherbench}.

These variables are the most prevalent commonality between different model developers' assessments; that is, they are used by default in reporting model skill for their meteorological importance as outlined above. In their original papers, Pangu-Weather \citep{bi2022pangu}, Spherical CNN \citep{esteves23sphericalcnn}, FourCastNet \citep{pathak2022fourcastnet}, FuXi \citep{chen2023fuxi}, Keisler \citep{keisler2022forecasting}, and NeuralGCM \citep{kochkov2024neuralgcm} are primarily evaluated with T850 and Z500, while GraphCast is an outlier \citep{lam2023learninggraphcast} reporting mainly on just Z500.

\subsection{Experimental design}

The main data we collect in our experiments involves taking the area-weighted squared difference between the models prediction $\hat{y}$ and the ERA5 ground truth value $y$ at every individual gridpoint, for every lead time $\tau\in$\{12h, 24h, ..., 240h\}, at every 12 hour interval in 2020. 

% For each $\tau$, we first get the RMSE by averaging over the spatial and temporal (but not lead time) dimensions. This serves as a baseline, and is the RMSE that would often get reported in the weather forecasting literature.

Then, for each of our four attributes and both variables, we calculate the per-strata RMSE (averaged temporally over the year) at all ten lead times by taking the RMSE when spatially averaging over only the gridpoints within that strata. This allows us to see which stratum the models are performing best or worst within.

Lastly, for each attribute and variable, we take the greatest absolute difference in per-strata RMSE of any pair of per-strata RMSE with the same attribute and variable. We also take the variance of all the per-strata RMSE to characterize the spread of model performance. This allows us to quantify the fairness of a model's predictions, where the smaller the difference and variance are, the more fair the model.

% Calculating these metrics for all six models took under 8 hours on a single CPU. At least 16GB of free storage is necessary to store intermediate data.

\subsection{Results}

\begin{figure}
  \centering
  \includegraphics[width=0.85\linewidth]{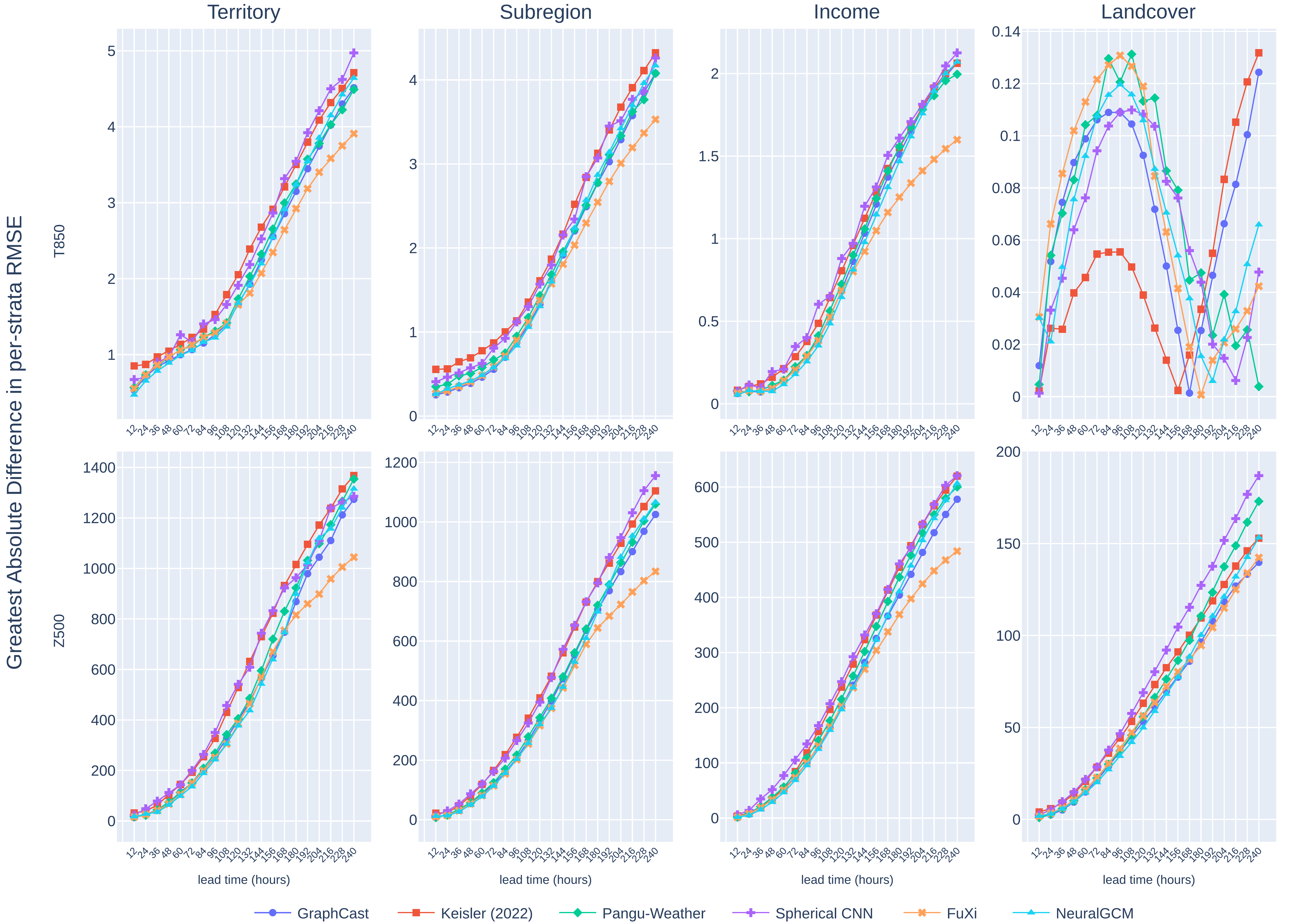}
  \caption{Greatest absolute difference of any two per-strata RMSE for each attribute when predicting T850 and Z500 at different lead times. Lower difference is more fair. Starting at a lead time of about one week, FuXi is the most fair model across all attributes and variables.}
  \label{rmse-diff}
\end{figure}

\begin{figure}
  \centering
  \includegraphics[width=0.85\linewidth]{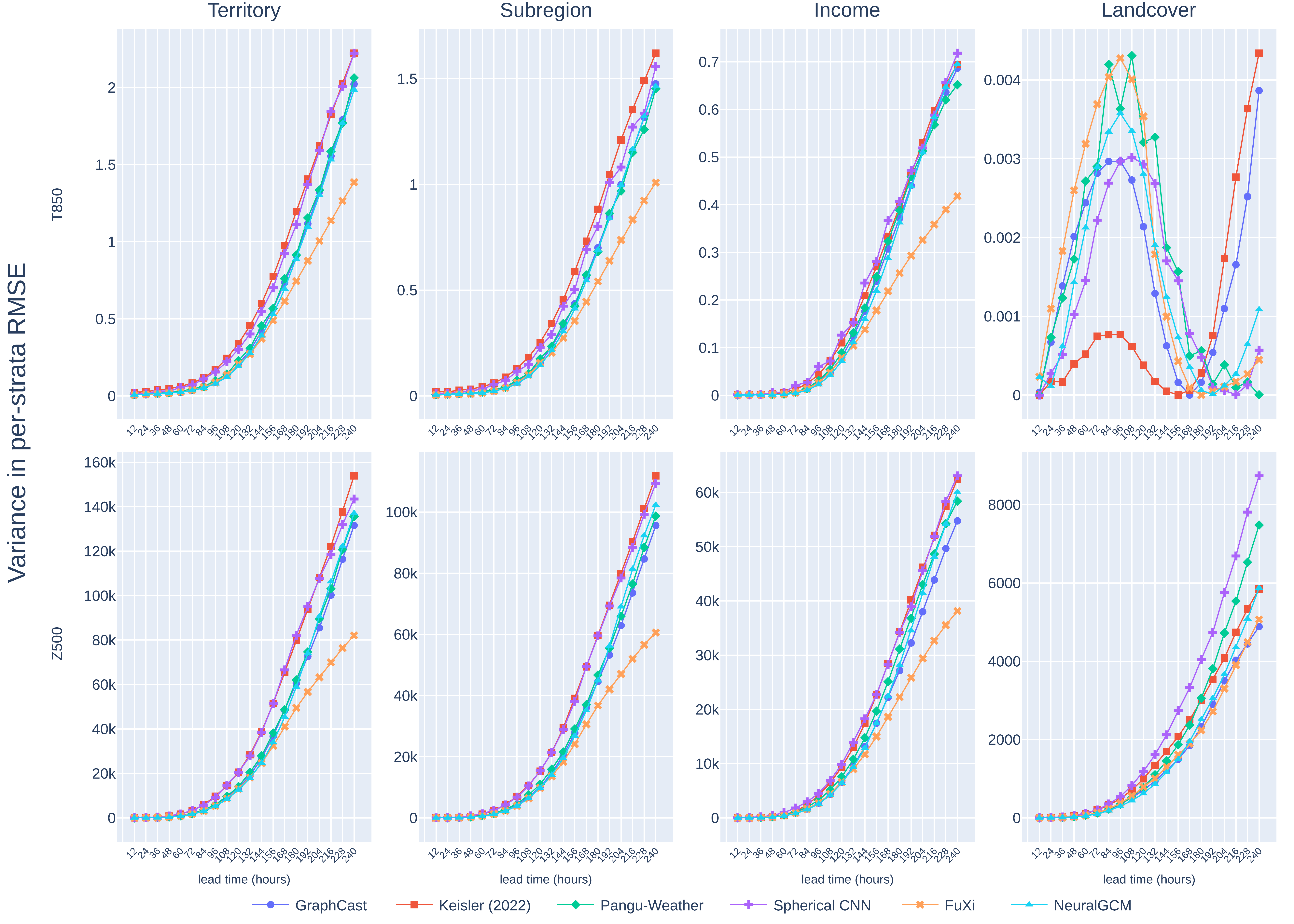}
  \caption{Variance of all the per-strata RMSE for each attribute when predicting T850 and Z500 at different lead times. Lower variance is more fair.}
  \label{rmse-var}
\end{figure}

\textbf{General fairness.} As seen in \autoref{rmse-diff}, the fairness of predictions begin to rapidly decline once the lead time surpasses three days; that is, the greatest absolute difference in RMSE of any two strata rapidly increases. Across all four attributes and all lead times, Spherical CNN and Keisler are generally the least fair. From a lead time of about a week onwards, FuXi is drastically more fair than every other model across all attributes. At early lead times, NeuralGCM appears to perform most fairly. We provide comprehensive benchmarks of the model fairness results in \autoref{fairness-benchmark-tables}. We also calculate the variance in per-strata RMSEs which displays similar patterns as seen in \autoref{rmse-var}. The main difference with variance is that it takes a larger lead time for unfairness to exponentially increase. \autoref{outliers} proves this discovered unfairness is not driven by outliers. 

% BETTER RESPONSE IS TAKEN IN 4.5 - THIS ACTUALLY GOES AGAINST OUR RESULTS: To dispel the notion that it may be extreme outliers that driving the unfairness results, we looked at the distribution for the attribute with the largest set of strata: territory. Across all six models, the distributions of territorial RMSEs have non-sparse long tails. For T850, the mean percent across all models and lead times of per-strata RMSEs outliers is 0.25\%. For Z500, it is 0.12\%. An outlier is defined as an RMSE either below the lower fence ($Q1-1.5*IQR$) or above the upper fence ($Q3+1.5*IQR$) of the distribution.

\begin{figure}
  \centering
  \includegraphics[width=0.85\linewidth]{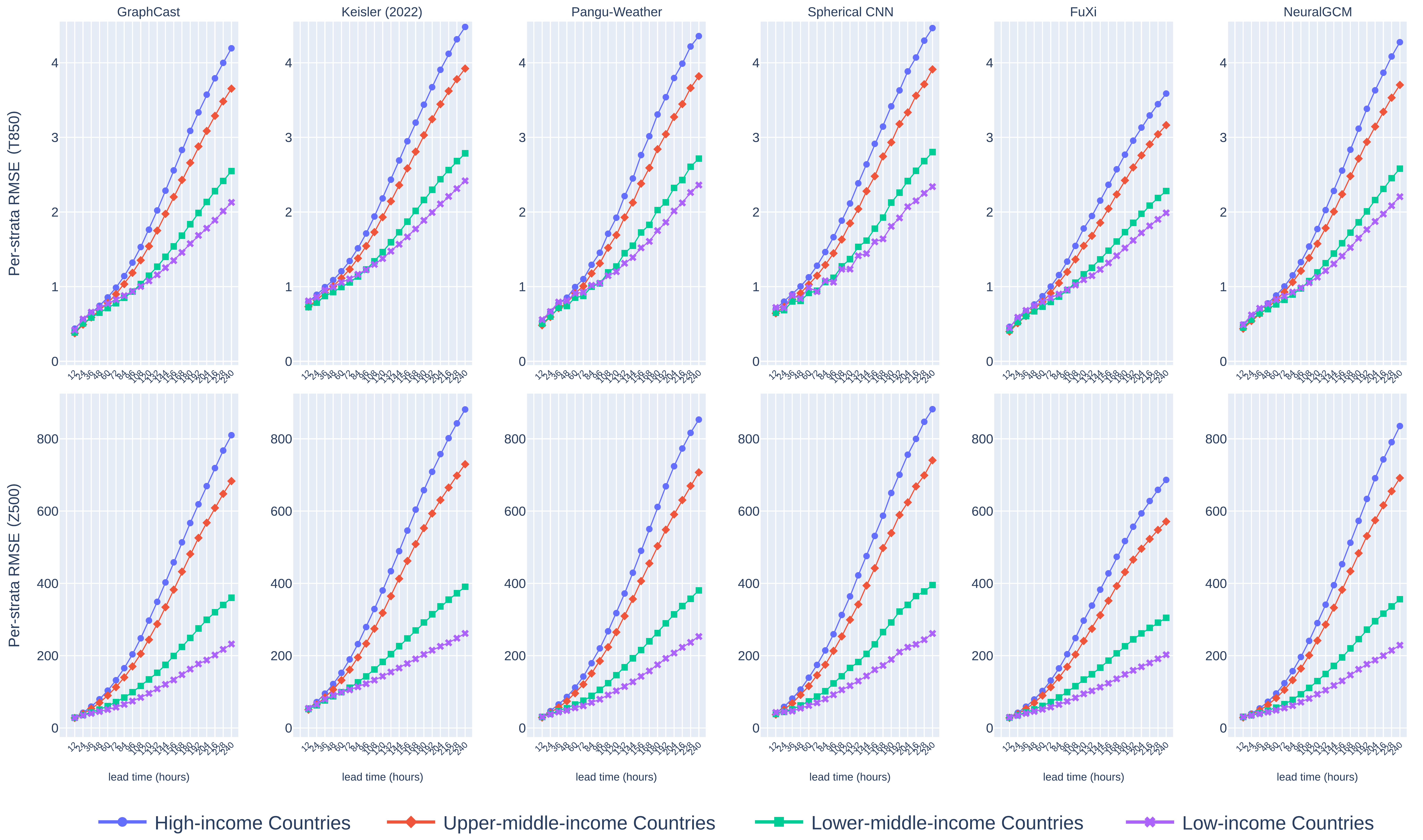}
  \caption{Per-strata RMSE for the income attribute of each model. This captures how well models perform at predicting each climatic variable stratified by the income classification for the associated country. We see that a bias against high income countries grows over time.}
  \label{fig-income-per-strata}
\end{figure}

\textbf{Income attribute}. To qualitatively characterize the growing unfairness observed in \autoref{rmse-diff}, we take a detailed look at the income attribute. Because it only has four strata, it is easy to visualize and meaningful to explore. For lead time $\tau=12$ hours, Keisler, Pangu-Weather, Spherical CNN, and NeuralGCM perform worst at predicting both variables in low-income territories (\autoref{fig-zoomed-income}).\footnote{The exception being NeuralGCM, where the per-strata RMSE on Z500 for lower-middle-income is 30.60187 versus low-income's 30.58936.} However, by $\tau=48$ hours, every model displays the trend for both variables where prediction skill decreases as income increases; this disparity continues to grow with lead time (\autoref{fig-income-per-strata}). This is an interesting result, and it shows that lead time is an important dimension to consider, because the disparity observed at one fixed lead time may not hold at another.

\begin{figure}
  \centering
  \includegraphics[width=0.85\linewidth]{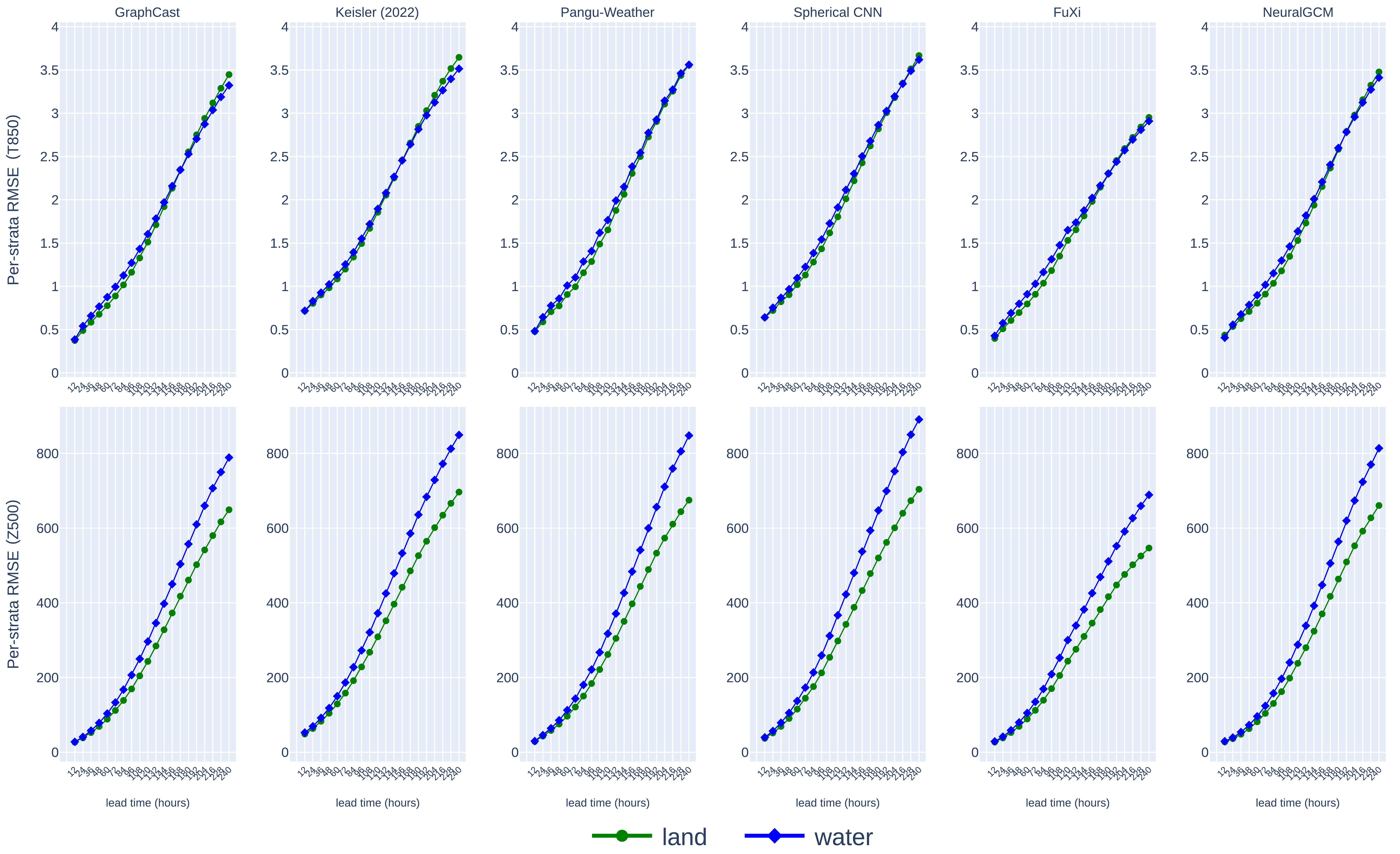}
  \caption{Per-strata RMSE for the landcover attribute of each model. This captures how well models perform at predicting each climatic variable stratified by the prediction being over land or water (oceans, seas, and many large lakes).}
  \label{fig-landcover-per-strata}
\end{figure}

\textbf{Landcover attribute.} We also take a close look at the landcover attribute. Generally, models perform better over land than water. This can be seen in \autoref{fig-landcover-per-strata}. However, by a lead time of 9 days ($\tau=216$ hours), all of the models except Pangu-Weather become worse at predicting temperature over land than water. In looking at greatest absolute difference and variance in RMSE, Pangu-Weather did not appear as the most fair with regards to landcover. However, we consider landcover to be a unique attribute. We have a special interest in absolute performance on the land stratum alone as that is where people live (small island nations are dutifully included in ``land'').\footnote{One exception is boats out at sea. For this case, \sys still provides state of the art advancements as model users can now look at model performance specifically on the oceanic gridpoints they will be traveling across.} In this sense, Pangu-Weather behaves as we may hope in always performing better over land than water, perhaps even more so than if it had equal performance across the strata. This is an exception to the fairness paradigm we laid out before, though it is sensible as all of the other attributes' strata are subsets of the ``land'' strata. In those cases, we want all the strata to be treated equally. Looking at the landcover attribute as a whole, FuXi is still the most overall fair as at given lead times it has the lowest error for the land stratum.

\subsection{Accounting for outliers} \label{outliers}

For each model we have assessed, the greatest absolute difference and variance in RMSE for each variable decreases as the number of stratum for the attribute decreases. This raises the question of whether the unfairness phenomenon observed results from rare outliers that appear as the geographic area of the smallest stratum decreases. To account for this, we reconduct our general fairness analysis after filtering out the set of outlier per-strata RMSE for every attribute. Because the data is skewed, we do not use Tukey's fences as a determination of outlyingness. Furthermore, as the data is bimodal at high lead times for the territory and subregion attributes, we cannot use the adjusted boxplot \citep{hubert2008adjusted} or adjusted outlyingness (AO) \citep{hubert2008outlier} methods either. Thus, we turn to local outlier factor (LOF) \citep{breunig2000lof} as our method of outlier detection. We use the default scikit-learn parameters.

\autoref{rmse-diff-no-outliers} and \autoref{rmse-var-no-outliers} are the same as \autoref{rmse-diff} and \autoref{rmse-var}, respectively, except the outliers have been filtered out. Because the landcover attribute only has two strata, the notion of an outlier does not make sense and so this attribute has been excluded. To more easily compare the results when both including and excluding outliers, we graph the largest per-strata RMSE as a percent of the smallest per-strata RMSE in \autoref{rmse-percent}. While there are slight differences in the greatest absolute difference in RMSE for the territory attribute (as evidenced by the different percentages), the general shape of the curves as a function of lead time holds with minor decreases in amplitude. This shows there are deeper trends in unfairness that are not being driven by outliers alone.

\begin{takeaway}
    An example analysis made possible through \textbf{\sys proves the existence of systemic bias in AI weather prediction} by location (at both territory and region resolution), income, and landcover for the climatic variables assessed.
\end{takeaway}

\section{Future work} \label{future-work}

An important future direction of work on improving \sys is incorporating more attributes. Moving beyond binary landcover, work with implicit neural representation (INR) models has shown that it is important to further consider coastlines and islands as their own strata as well \citep{cai2025no}. Additionally, population density will be added to \sys as an attribute to better understand the degree to which different AIWP models can be a trusted decision-making tools across different human settlements. This will improve on the already state-of-the-art territory-level precision of this work.

Currently, \sys operates at inference time. It may prove beneficial to integrate tracking of fairness metrics into the training regimes of models to understand how different training dynamics affect fairness. Further, incorporating spatial stratification into training objectives could ameliorate bias. In general, investigating the underlying causes for why different models are more or less fair and how to remedy this are consequential research questions that are first raised by our work.

% \section{Limitations} \label{limitations}

% The $1.5^\circ$ resolution used within this paper was beneficial in providing a common resolution to maximize how many models we could benchmark. However, higher resolution ameliorates the double counting issue described in \autoref{stratum-membership-methods}, providing more precise stratification. Important future steps involve reproducing the experiments in this work with model forecasts at higher spatial resolution. Importantly, such work is made possible for the first time by the introduction of the \sys package.

\section{Discussion}

Organizations like the NOAA are beginning to incorporate ML systems in their work, citing improvements in models such as ECMWF's very own Artificial Intelligence/Integrated Forecasting System (AIFS) \citep{nextgovnoaa}. As AIWP models become increasingly relied upon, the results of this work necessitates more careful attention being paid to the stratified performance and fairness of models. By using \sys to investigate the territory attribute, one is able to find whether a given AIWP is appropriate to leverage in decision making within that territory. This is an important discovery given the life and death consequences that forecasts can impart. The benchmark provided in this work is a first step in this direction. Moreover, \sys empowers deployers to select the model which is most performant for their local application given the biases we prove exist. The visibility provided by \sys into stratified forecast fairness brings this research area to light.

\section{Conclusion}

In this work we created \sys, a python package that allows the user to assess a set of machine learning predictions made over Earth in terms of stratified fairness. Strata are available for four attributes a gridpoint may have: territorial affiliation, global subregion, gross national income per capita, and landcover. This provides developers and decision-makers alike with an important tool to break free from the default approach of spatial averaging. We apply \sys to a set of state of the art AIWP models, finding that they all display unfair spatial disparities in performance on all four attributes. These disparities generally increase with lead time, particularly after three days. These findings justify our approach of capturing more geographically fine-tuned errors, discouraging the current reliance on spatially-averaged RMSE for characterizing AIWP models. This is an advancement upon the foundation of all AI weather and climate work.

% Through investigating the income attribute specifically, we found that there are systemic inequalities in AIWP forecast skill across different strata. Notably, models generally perform worse on low-income territories at the shortest lead times, followed by this trend completely reversing at around 2 days, where model forecasts display worse performance as income strata increases. In terms of geography, there are vast disparities in performance across fine-grain territories and coarse (but precise) global subregions. Lastly, models perform better over the ocean and sea than land. The global spatial averaging of models’ training objective may be contributing to them overfitting on the majority of Earth’s surface that is water, sacrificing performance over land—land where people live and where weather predictions are most important.

% TODO: uncomment section below
\subsubsection*{Acknowledgments}

The authors thank Daniel Cai and Philip LaDuca. Part of this research was conducted using computational resources and services at the Center for Computation and Visualization, Brown University. LLMs were used in debugging package code. No LLM was used in the writing of this paper.

\subsubsection*{Reproducibility statement}

We have made \sys open source, including code for reproducing the specific results of \autoref{experiments}, as well as the entire generic framework for promoting similar stratified evaluations on more models and datasets. Code for generating the figures of this work is also included in the repo. We have stylistically altered them in a vector graphics editor, but the data values and representation are the same as those output by the scripts in \mintinline{bash}{demos/} directory. We clearly state the origin of our climate and attribute data in the main text of the paper in \autoref{data-sources} and \autoref{experimental-details}; we also go into further detail in \autoref{attribute-strata-details}. The code for generating all of the data is part of the \sys repo in \mintinline{bash}{src/safe_earth/data/}. In \autoref{safe-metrics}, we provide code snippets for calculating the metrics we report on.

\bibliography{bib}
\bibliographystyle{iclr2026_conference}

\appendix

% \section{LLM usage} \label{llm-usage}

% LLMs were used in debugging package code. No LLM was used in the writing of this paper.

\section{Model details} \label{model-details}

Original papers, architecture type, and number of parameters (if reported in the original paper) are described in \autoref{models-list} for all of the models we asses in our demonstration of \sys in \autoref{experiments}.

\begin{table}[!htb]
  \caption{Models assessed}
  \label{models-list}
  \scriptsize
  \centering
  \begin{tabular}{p{.225\linewidth}p{.525\linewidth}p{.1\linewidth}}
    \\ \toprule
    Model & Architecture & Parameters \\
    \midrule
    GraphCast \citep{lam2023learninggraphcast} & Graph neural network (GNN) & 36.7 M \\
    Keisler \citep{keisler2022forecasting} & GNN & 6.7 M \\
    Pangu-Weather \citep{bi2022pangu} & Earth-specific transformer & 256 M \\
    Spherical CNN \citep{esteves23sphericalcnn} & Spherical convolutional neural network (CNN) & Not reported \\
    FuXi \citep{chen2023fuxi} & SwinV2 \citep{liu2022swinv2} transformer blocks in U-net \citep{ronneberger2015unet} arrangement & Not reported \\
    NeuralGCM \citep{kochkov2024neuralgcm} & Multi-layer perceptrons (MLP) + CNNs + numerical solver & 31.1 M \\
    \bottomrule
  \end{tabular}
\end{table}

\section{Attribute strata details}  \label{attribute-strata-details}

Interactive maps showing the geographic locations of each strata for the different attributes is available at \url{https://n-masi.github.io/safe}.

\subsection{Subregions}

The 23 strata included in the global subregion attribute are: Antarctica, Australia/New Zealand, Caribbean, Central America, Central Asia, Eastern Africa, Eastern Asia, Eastern Europe, Melanesia, Micronesia, Middle Africa, Northern Africa, Northern America, Northern Europe, Polynesia, South America, South-Eastern Asia, Southern Africa, Southern Asia, Southern Europe, Western Africa, Western Asia, and Western Europe.

\subsection{Income}

The 18 territory strata without income classifications by the World Bank are: Anguilla; Antarctica; Bonaire, Sint Eustatius, and Saba; Saint Barthelemy; Cook Islands; Falkland Islands; Guadeloupe; French Guiana; Montserrat; Martinique; Mayotte; Niue; Pitcairn Island; Réunion; Saint Helena, Ascension, and Tristan da Cunha; Vatican City; Wallis and Futuna; and Tokelau.

\subsection{Landcover}

Due to idiosyncrasies in geoBoundaries, the landcover strata of ``land'' includes most lakes. Of the 15 largest by surface area, the following are included in ``land'': Lake Baikal, Great Bear Lake, Great Slave Lake, Lake Winnipeg, Lake Ladoga, and Lake Balkhash. Disambiguation will be improved through the integration of data sources more targeted for landcover. Datasets such as LandScan Global \citep{dobson2000landscan, lebakula2024landscan} can provide gridpoint information with the strata of landmass, ocean, lake, land within lake, and so on.

\section{Fairness Benchmarks} \label{fairness-benchmark-tables}

\begin{table}[H]
	\caption{Greatest absolute difference in per-strata RMSE for territory. Lower is more fair; most fair for each variable and lead time is bolded.}
	\label{territory-gad_rmse_weighted_l2-benchmark}
	\scriptsize
	\centering
	\begin{tabular}{lllllllll}
		\\ \toprule
		& & \multicolumn{6}{c}{Model} \\
		\cmidrule(r){3-8}
		Variable & Lead time (h) & GraphCast & Keisler & Pangu-Weather & Spherical CNN & FuXi & NeuralGCM \\
		\midrule
		T850 & 12h & 0.5301 & 0.8523 & 0.5677 & 0.6726 & 0.5548 & \textbf{0.4715} \\
		T850 & 24h & 0.7129 & 0.8712 & 0.7346 & 0.7011 & 0.7321 & \textbf{0.6562} \\
		T850 & 36h & 0.8428 & 0.9704 & 0.8578 & 0.9009 & 0.8646 & \textbf{0.7861} \\
		T850 & 48h & 0.9265 & 1.0467 & 0.9562 & 0.9670 & 0.9731 & \textbf{0.8921} \\
		T850 & 60h & 0.9991 & 1.1356 & 1.0671 & 1.2620 & 1.0586 & \textbf{0.9915} \\
		T850 & 72h & 1.0666 & 1.2268 & 1.1138 & 1.1681 & 1.1301 & \textbf{1.0552} \\
		T850 & 84h & \textbf{1.1528} & 1.3355 & 1.2301 & 1.4047 & 1.2289 & 1.1660 \\
		T850 & 96h & 1.2629 & 1.5298 & 1.3096 & 1.4621 & 1.2870 & \textbf{1.2227} \\
		T850 & 108h & 1.3920 & 1.7896 & 1.4194 & 1.6604 & 1.4135 & \textbf{1.3685} \\
		T850 & 120h & 1.6689 & 2.0537 & 1.7343 & 1.9129 & \textbf{1.6558} & 1.6819 \\
		T850 & 132h & 1.9286 & 2.3906 & 2.0301 & 2.1865 & \textbf{1.8117} & 1.9187 \\
		T850 & 144h & 2.2473 & 2.6782 & 2.3223 & 2.5244 & \textbf{2.0726} & 2.1994 \\
		T850 & 156h & 2.5549 & 2.9147 & 2.6536 & 2.8648 & \textbf{2.3466} & 2.5437 \\
		T850 & 168h & 2.8560 & 3.2085 & 2.9975 & 3.3171 & \textbf{2.6423} & 2.9141 \\
		T850 & 180h & 3.1494 & 3.5049 & 3.2451 & 3.5449 & \textbf{2.9232} & 3.2373 \\
		T850 & 192h & 3.4473 & 3.7977 & 3.5734 & 3.9219 & \textbf{3.1848} & 3.5434 \\
		T850 & 204h & 3.7448 & 4.0875 & 3.7818 & 4.2116 & \textbf{3.4015} & 3.8543 \\
		T850 & 216h & 4.0216 & 4.3177 & 4.0281 & 4.4993 & \textbf{3.5835} & 4.1497 \\
		T850 & 228h & 4.2999 & 4.5044 & 4.2233 & 4.6229 & \textbf{3.7494} & 4.4231 \\
		T850 & 240h & 4.5130 & 4.7116 & 4.4912 & 4.9728 & \textbf{3.9086} & 4.6413 \\
		\midrule
		Z500 & 12h & \textbf{13.4222} & 31.7980 & 17.1554 & 23.3231 & 15.6101 & 17.7155 \\
		Z500 & 24h & 25.4911 & 39.2029 & \textbf{22.7217} & 48.5697 & 25.4045 & 27.7922 \\
		Z500 & 36h & 44.5530 & 63.3959 & 44.6666 & 78.6476 & 41.8652 & \textbf{35.5452} \\
		Z500 & 48h & 73.7853 & 101.7007 & 74.2397 & 112.7660 & 67.0475 & \textbf{63.8041} \\
		Z500 & 60h & 114.5105 & 144.6675 & 112.3888 & 143.2511 & 104.6706 & \textbf{98.8060} \\
		Z500 & 72h & 149.7483 & 192.5330 & 150.8271 & 199.7432 & 150.0260 & \textbf{136.3233} \\
		Z500 & 84h & 197.7661 & 254.1886 & 209.0000 & 263.5604 & 199.1568 & \textbf{189.1055} \\
		Z500 & 96h & 258.1191 & 326.9143 & 268.7369 & 350.5695 & 251.8137 & \textbf{243.2503} \\
		Z500 & 108h & 327.6491 & 430.2865 & 341.8734 & 457.4359 & \textbf{305.2133} & 305.4085 \\
		Z500 & 120h & 396.9567 & 528.5107 & 405.2052 & 541.2818 & 386.7127 & \textbf{377.4789} \\
		Z500 & 132h & 471.5472 & 631.6456 & 485.4492 & 608.7734 & 464.0384 & \textbf{437.4839} \\
		Z500 & 144h & 566.6680 & 729.6151 & 595.1192 & 743.3179 & 569.2445 & \textbf{542.7397} \\
		Z500 & 156h & 654.4902 & 822.9236 & 720.2494 & 832.6830 & 669.9283 & \textbf{639.2373} \\
		Z500 & 168h & 747.2870 & 932.4173 & 830.2883 & 922.7379 & 754.8806 & \textbf{747.1750} \\
		Z500 & 180h & 868.6595 & 1015.7405 & 923.0644 & 963.3164 & \textbf{815.4693} & 898.6278 \\
		Z500 & 192h & 979.1869 & 1095.5909 & 1031.9875 & 1014.1516 & \textbf{859.9232} & 1023.4053 \\
		Z500 & 204h & 1044.4335 & 1171.7235 & 1098.2249 & 1110.0590 & \textbf{898.6318} & 1120.0131 \\
		Z500 & 216h & 1110.4524 & 1237.9779 & 1173.1740 & 1239.1351 & \textbf{958.4204} & 1156.8152 \\
		Z500 & 228h & 1212.0978 & 1314.9546 & 1266.0661 & 1261.9217 & \textbf{1005.6409} & 1239.9979 \\
		Z500 & 240h & 1273.6925 & 1367.5504 & 1354.0935 & 1285.5398 & \textbf{1044.5271} & 1315.3407 \\
		\bottomrule
	\end{tabular}
\end{table}

\begin{table}[H]
	\caption{Greatest absolute difference in per-strata RMSE for global subregion. Lower is more fair; most fair for each variable and lead time is bolded.}
	\label{subregion-gad_rmse_weighted_l2-benchmark}
	\scriptsize
	\centering
	\begin{tabular}{lllllllll}
		\\ \toprule
		& & \multicolumn{6}{c}{Model} \\
		\cmidrule(r){3-8}
		Variable & Lead time (h) & GraphCast & Keisler & Pangu-Weather & Spherical CNN & FuXi & NeuralGCM \\
		\midrule
		T850 & 12h & \textbf{0.2525} & 0.5555 & 0.3504 & 0.4085 & 0.2690 & 0.2599 \\
		T850 & 24h & \textbf{0.2852} & 0.5617 & 0.3779 & 0.4630 & 0.2981 & 0.3243 \\
		T850 & 36h & \textbf{0.3363} & 0.6454 & 0.4780 & 0.5130 & 0.3512 & 0.3723 \\
		T850 & 48h & \textbf{0.3880} & 0.6931 & 0.5075 & 0.5734 & 0.4074 & 0.4177 \\
		T850 & 60h & \textbf{0.4623} & 0.7767 & 0.5796 & 0.6264 & 0.4834 & 0.4862 \\
		T850 & 72h & \textbf{0.5562} & 0.8714 & 0.6702 & 0.8082 & 0.5863 & 0.5778 \\
		T850 & 84h & 0.6994 & 1.0025 & 0.7491 & 0.9256 & 0.7134 & \textbf{0.6865} \\
		T850 & 96h & 0.8757 & 1.1339 & 0.9501 & 1.1201 & 0.9017 & \textbf{0.8381} \\
		T850 & 108h & 1.0833 & 1.3553 & 1.1729 & 1.3028 & 1.1138 & \textbf{1.0592} \\
		T850 & 120h & 1.3296 & 1.6098 & 1.4336 & 1.5700 & 1.3764 & \textbf{1.3057} \\
		T850 & 132h & 1.6137 & 1.8671 & 1.6829 & 1.7964 & \textbf{1.5735} & 1.5992 \\
		T850 & 144h & 1.9168 & 2.1653 & 1.9563 & 2.1541 & \textbf{1.8085} & 1.9220 \\
		T850 & 156h & 2.2034 & 2.5199 & 2.2203 & 2.3437 & \textbf{2.0344} & 2.2378 \\
		T850 & 168h & 2.4912 & 2.8395 & 2.5085 & 2.8486 & \textbf{2.2963} & 2.5705 \\
		T850 & 180h & 2.7716 & 3.1276 & 2.7775 & 3.0715 & \textbf{2.5446} & 2.8682 \\
		T850 & 192h & 3.0267 & 3.4037 & 3.1109 & 3.4501 & \textbf{2.7923} & 3.1401 \\
		T850 & 204h & 3.2891 & 3.6766 & 3.3345 & 3.5138 & \textbf{3.0081} & 3.4233 \\
		T850 & 216h & 3.5752 & 3.9070 & 3.6220 & 3.7692 & \textbf{3.1939} & 3.6968 \\
		T850 & 228h & 3.8532 & 4.1115 & 3.7643 & 3.8652 & \textbf{3.3680} & 3.9615 \\
		T850 & 240h & 4.0787 & 4.3223 & 4.0785 & 4.2605 & \textbf{3.5287} & 4.1710 \\
		\midrule
		Z500 & 12h & 10.4583 & 22.0202 & \textbf{7.0142} & 13.0860 & 9.6233 & 12.3408 \\
		Z500 & 24h & 15.2147 & 23.3950 & 15.3775 & 29.7619 & 14.1225 & \textbf{13.2607} \\
		Z500 & 36h & 31.6671 & 47.2062 & 35.2885 & 52.6038 & 30.2638 & \textbf{27.0468} \\
		Z500 & 48h & 55.0884 & 79.2753 & 56.8891 & 87.0013 & 53.2712 & \textbf{50.2599} \\
		Z500 & 60h & 84.5594 & 118.5651 & 89.9188 & 120.1655 & 81.1725 & \textbf{77.4975} \\
		Z500 & 72h & 119.6162 & 165.2460 & 124.3231 & 161.8944 & 114.0495 & \textbf{113.2731} \\
		Z500 & 84h & 159.4383 & 218.3120 & 169.9461 & 206.6974 & \textbf{154.0163} & 156.4037 \\
		Z500 & 96h & 206.9141 & 276.7293 & 217.2961 & 265.7229 & \textbf{201.8156} & 203.6802 \\
		Z500 & 108h & 265.2308 & 340.8177 & 278.5598 & 324.6950 & \textbf{254.1852} & 257.5620 \\
		Z500 & 120h & 330.6048 & 409.0917 & 342.7813 & 394.7158 & \textbf{315.9001} & 320.4175 \\
		Z500 & 132h & 398.4901 & 481.7423 & 408.0717 & 477.3002 & \textbf{374.1447} & 375.0295 \\
		Z500 & 144h & 472.8915 & 560.7308 & 480.2798 & 572.4723 & \textbf{442.6990} & 445.9015 \\
		Z500 & 156h & 553.3959 & 647.4817 & 560.9786 & 652.9538 & \textbf{519.2081} & 530.1986 \\
		Z500 & 168h & 635.8493 & 731.0427 & 640.5657 & 732.2401 & \textbf{589.3926} & 609.7468 \\
		Z500 & 180h & 705.7083 & 800.0373 & 720.1483 & 795.3998 & \textbf{644.0070} & 698.6875 \\
		Z500 & 192h & 769.0193 & 861.8439 & 789.5368 & 880.8382 & \textbf{684.0959} & 791.0014 \\
		Z500 & 204h & 833.2273 & 928.2371 & 863.7943 & 947.5966 & \textbf{722.9894} & 882.7328 \\
		Z500 & 216h & 900.5285 & 993.3099 & 932.0408 & 1031.0736 & \textbf{765.0671} & 953.4226 \\
		Z500 & 228h & 968.6828 & 1051.8553 & 1003.9197 & 1105.4598 & \textbf{802.7399} & 1011.1210 \\
		Z500 & 240h & 1025.1756 & 1104.4001 & 1060.0999 & 1155.7330 & \textbf{833.8870} & 1066.3462 \\
		\bottomrule
	\end{tabular}
\end{table}

\begin{table}[H]
	\caption{Greatest absolute difference in per-strata RMSE for income. Lower is more fair; most fair for each variable and lead time is bolded.}
	\label{income-gad_rmse_weighted_l2-benchmark}
	\scriptsize
	\centering
	\begin{tabular}{lllllllll}
		\\ \toprule
		& & \multicolumn{6}{c}{Model} \\
		\cmidrule(r){3-8}
		Variable & Lead time (h) & GraphCast & Keisler & Pangu-Weather & Spherical CNN & FuXi & NeuralGCM \\
		\midrule
		T850 & 12h & 0.0620 & 0.0825 & 0.0751 & 0.0774 & 0.0642 & \textbf{0.0542} \\
		T850 & 24h & 0.0754 & 0.1048 & \textbf{0.0727} & 0.1140 & 0.0797 & 0.0805 \\
		T850 & 36h & \textbf{0.0715} & 0.1201 & 0.0813 & 0.0972 & 0.0757 & 0.0758 \\
		T850 & 48h & 0.0937 & 0.1607 & 0.1106 & 0.1952 & 0.0928 & \textbf{0.0754} \\
		T850 & 60h & 0.1423 & 0.2117 & 0.1430 & 0.2091 & 0.1405 & \textbf{0.1181} \\
		T850 & 72h & 0.2078 & 0.2854 & 0.2241 & 0.3468 & 0.2050 & \textbf{0.1791} \\
		T850 & 84h & 0.2900 & 0.3773 & 0.2926 & 0.4009 & 0.2881 & \textbf{0.2559} \\
		T850 & 96h & 0.3875 & 0.4873 & 0.4117 & 0.6024 & 0.3822 & \textbf{0.3522} \\
		T850 & 108h & 0.5282 & 0.6422 & 0.5619 & 0.6514 & 0.5243 & \textbf{0.4859} \\
		T850 & 120h & 0.6864 & 0.8056 & 0.7235 & 0.8795 & 0.6860 & \textbf{0.6450} \\
		T850 & 132h & 0.8621 & 0.9584 & 0.8995 & 0.9707 & \textbf{0.8004} & 0.8119 \\
		T850 & 144h & 1.0326 & 1.1233 & 1.0585 & 1.1959 & \textbf{0.9227} & 0.9777 \\
		T850 & 156h & 1.2090 & 1.2809 & 1.2427 & 1.3127 & \textbf{1.0478} & 1.1458 \\
		T850 & 168h & 1.3726 & 1.4249 & 1.4092 & 1.5043 & \textbf{1.1584} & 1.3107 \\
		T850 & 180h & 1.5107 & 1.5502 & 1.5566 & 1.6087 & \textbf{1.2506} & 1.4676 \\
		T850 & 192h & 1.6483 & 1.6787 & 1.6764 & 1.7082 & \textbf{1.3365} & 1.6183 \\
		T850 & 204h & 1.7905 & 1.7966 & 1.7819 & 1.8130 & \textbf{1.4103} & 1.7575 \\
		T850 & 216h & 1.9016 & 1.9114 & 1.8654 & 1.9200 & \textbf{1.4795} & 1.8932 \\
		T850 & 228h & 1.9866 & 2.0008 & 1.9558 & 2.0456 & \textbf{1.5437} & 1.9994 \\
		T850 & 240h & 2.0647 & 2.0616 & 1.9952 & 2.1247 & \textbf{1.5983} & 2.0702 \\
		\midrule
		Z500 & 12h & \textbf{0.8108} & 3.6642 & 1.6727 & 5.9048 & 1.5137 & 1.6957 \\
		Z500 & 24h & 7.2642 & 8.9651 & 8.9447 & 13.8836 & 7.8770 & \textbf{5.0367} \\
		Z500 & 36h & 18.9145 & 19.0145 & 20.9362 & 34.6394 & 18.7908 & \textbf{15.1110} \\
		Z500 & 48h & 34.0168 & 34.0692 & 37.4726 & 51.7026 & 33.1263 & \textbf{28.8815} \\
		Z500 & 60h & 52.3629 & 53.6224 & 56.2572 & 77.0541 & 50.8687 & \textbf{46.4533} \\
		Z500 & 72h & 74.6146 & 84.2772 & 80.9830 & 104.9152 & 73.3257 & \textbf{68.6393} \\
		Z500 & 84h & 100.2624 & 118.3149 & 108.8336 & 134.4815 & 100.0320 & \textbf{95.4571} \\
		Z500 & 96h & 129.4711 & 156.8515 & 140.8503 & 167.4550 & 130.5978 & \textbf{124.9093} \\
		Z500 & 108h & 163.7621 & 196.7506 & 176.3627 & 207.2879 & 165.2451 & \textbf{159.4563} \\
		Z500 & 120h & 201.4575 & 237.4061 & 215.1524 & 247.2022 & 202.5977 & \textbf{196.7490} \\
		Z500 & 132h & 240.7642 & 279.0208 & 257.4120 & 292.4811 & \textbf{236.1156} & 236.5664 \\
		Z500 & 144h & 282.1859 & 323.2796 & 301.8338 & 331.7819 & \textbf{269.7027} & 277.7673 \\
		Z500 & 156h & 325.4654 & 367.9642 & 347.4286 & 370.2875 & \textbf{304.1304} & 323.0253 \\
		Z500 & 168h & 366.2331 & 413.1306 & 392.5974 & 414.4043 & \textbf{337.6676} & 365.8929 \\
		Z500 & 180h & 404.1501 & 454.5309 & 436.6129 & 460.6266 & \textbf{368.8637} & 410.7149 \\
		Z500 & 192h & 441.7336 & 493.7207 & 476.3061 & 490.5075 & \textbf{397.4898} & 457.3617 \\
		Z500 & 204h & 481.4574 & 531.8207 & 516.7449 & 532.7525 & \textbf{424.5340} & 503.4393 \\
		Z500 & 216h & 517.2955 & 566.0298 & 550.1209 & 568.5136 & \textbf{448.0975} & 543.1820 \\
		Z500 & 228h & 550.2756 & 594.4706 & 579.6324 & 602.9630 & \textbf{467.4228} & 576.0455 \\
		Z500 & 240h & 577.7541 & 619.7738 & 600.2285 & 620.6610 & \textbf{483.7225} & 606.3814 \\
		\bottomrule
	\end{tabular}
\end{table}

\begin{table}[H]
	\caption{Greatest absolute difference in per-strata RMSE for landcover. Lower is more fair; most fair for each variable and lead time is bolded.}
	\label{landcover-gad_rmse_weighted_l2-benchmark}
	\scriptsize
	\centering
	\begin{tabular}{lllllllll}
		\\ \toprule
		& & \multicolumn{6}{c}{Model} \\
		\cmidrule(r){3-8}
		Variable & Lead time (h) & GraphCast & Keisler & Pangu-Weather & Spherical CNN & FuXi & NeuralGCM \\
		\midrule
		T850 & 12h & 0.0119 & 0.0022 & 0.0047 & \textbf{0.0014} & 0.0305 & 0.0301 \\
		T850 & 24h & 0.0518 & 0.0262 & 0.0542 & 0.0331 & 0.0662 & \textbf{0.0211} \\
		T850 & 36h & 0.0744 & \textbf{0.0258} & 0.0703 & 0.0454 & 0.0855 & 0.0497 \\
		T850 & 48h & 0.0897 & \textbf{0.0398} & 0.0831 & 0.0640 & 0.1020 & 0.0756 \\
		T850 & 60h & 0.0988 & \textbf{0.0456} & 0.1042 & 0.0762 & 0.1130 & 0.0922 \\
		T850 & 72h & 0.1061 & \textbf{0.0546} & 0.1077 & 0.0942 & 0.1215 & 0.1076 \\
		T850 & 84h & 0.1089 & \textbf{0.0554} & 0.1296 & 0.1037 & 0.1271 & 0.1156 \\
		T850 & 96h & 0.1090 & \textbf{0.0555} & 0.1206 & 0.1090 & 0.1308 & 0.1196 \\
		T850 & 108h & 0.1045 & \textbf{0.0497} & 0.1313 & 0.1099 & 0.1266 & 0.1158 \\
		T850 & 120h & 0.0925 & \textbf{0.0389} & 0.1132 & 0.1083 & 0.1189 & 0.1059 \\
		T850 & 132h & 0.0718 & \textbf{0.0263} & 0.1144 & 0.1036 & 0.0845 & 0.0873 \\
		T850 & 144h & 0.0500 & \textbf{0.0140} & 0.0865 & 0.0825 & 0.0631 & 0.0705 \\
		T850 & 156h & 0.0254 & \textbf{0.0024} & 0.0791 & 0.0762 & 0.0415 & 0.0541 \\
		T850 & 168h & \textbf{0.0014} & 0.0160 & 0.0446 & 0.0560 & 0.0190 & 0.0376 \\
		T850 & 180h & 0.0253 & 0.0335 & 0.0475 & 0.0439 & \textbf{0.0007} & 0.0155 \\
		T850 & 192h & 0.0465 & 0.0549 & 0.0235 & 0.0201 & 0.0139 & \textbf{0.0059} \\
		T850 & 204h & 0.0663 & 0.0833 & 0.0392 & \textbf{0.0147} & 0.0206 & 0.0219 \\
		T850 & 216h & 0.0814 & 0.1052 & 0.0195 & \textbf{0.0062} & 0.0259 & 0.0327 \\
		T850 & 228h & 0.1004 & 0.1206 & 0.0256 & \textbf{0.0227} & 0.0328 & 0.0507 \\
		T850 & 240h & 0.1243 & 0.1318 & \textbf{0.0039} & 0.0478 & 0.0423 & 0.0659 \\
		\midrule
		Z500 & 12h & 1.1498 & 4.0162 & \textbf{0.9773} & 2.4792 & 1.6285 & 1.8373 \\
		Z500 & 24h & \textbf{2.5139} & 5.8507 & 2.7293 & 5.1420 & 3.3727 & 2.9665 \\
		Z500 & 36h & \textbf{5.1433} & 8.9834 & 5.8332 & 9.6810 & 6.1696 & 5.7374 \\
		Z500 & 48h & \textbf{9.4250} & 13.8543 & 10.1021 & 14.7946 & 10.4900 & 9.5702 \\
		Z500 & 60h & 14.8999 & 20.7261 & 16.1918 & 21.7804 & 16.0543 & \textbf{14.4554} \\
		Z500 & 72h & 21.4036 & 28.3228 & 22.4379 & 28.5432 & 22.7172 & \textbf{20.1387} \\
		Z500 & 84h & 28.8944 & 36.0126 & 29.9796 & 37.6959 & 30.2328 & \textbf{27.1628} \\
		Z500 & 96h & 37.2341 & 44.2808 & 37.3130 & 46.6026 & 38.5729 & \textbf{34.4391} \\
		Z500 & 108h & 45.4507 & 53.2869 & 45.8862 & 57.6306 & 47.2646 & \textbf{41.9945} \\
		Z500 & 120h & 53.2044 & 63.1831 & 55.7587 & 68.9327 & 56.2924 & \textbf{49.9061} \\
		Z500 & 132h & 61.3276 & 73.3235 & 66.3274 & 80.2869 & 63.5645 & \textbf{58.8711} \\
		Z500 & 144h & 69.6294 & 82.5196 & 76.2810 & 92.0807 & 72.1485 & \textbf{68.2075} \\
		Z500 & 156h & 77.3385 & 91.0754 & 86.4044 & 104.6094 & 80.2376 & \textbf{77.3269} \\
		Z500 & 168h & \textbf{85.9806} & 100.1707 & 97.3472 & 115.3360 & 87.0638 & 88.4507 \\
		Z500 & 180h & 96.4382 & 109.5743 & 110.5671 & 127.2699 & \textbf{94.6042} & 100.2435 \\
		Z500 & 192h & 107.7786 & 118.8574 & 123.4246 & 137.6266 & \textbf{104.2935} & 110.5522 \\
		Z500 & 204h & 118.3307 & 127.7590 & 137.4269 & 151.7028 & \textbf{114.9430} & 121.0782 \\
		Z500 & 216h & 126.9011 & 137.7304 & 148.8532 & 163.5775 & \textbf{124.9670} & 131.9841 \\
		Z500 & 228h & \textbf{133.3081} & 146.0886 & 161.5919 & 176.7622 & 133.9938 & 142.6224 \\
		Z500 & 240h & \textbf{139.7696} & 152.9192 & 172.9741 & 186.9395 & 142.3599 & 153.2736 \\
		\bottomrule
	\end{tabular}
\end{table}

\begin{table}[H]
	\caption{Variance of per-strata RMSE for territory. Lower is more fair; most fair for each variable and lead time is bolded.}
	\label{territory-variance_rmse_weighted_l2-benchmark}
	\scriptsize
	\centering
	\begin{tabular}{lllllllll}
		\\ \toprule
		& & \multicolumn{6}{c}{Model} \\
		\cmidrule(r){3-8}
		Variable & Lead time (h) & GraphCast & Keisler & Pangu-Weather & Spherical CNN & FuXi & NeuralGCM \\
		\midrule
		T850 & 12h & 0.0059 & 0.0239 & 0.0092 & 0.0165 & \textbf{0.0058} & 0.0076 \\
		T850 & 24h & \textbf{0.0096} & 0.0279 & 0.0121 & 0.0168 & 0.0097 & 0.0107 \\
		T850 & 36h & \textbf{0.0135} & 0.0382 & 0.0186 & 0.0307 & 0.0138 & 0.0142 \\
		T850 & 48h & \textbf{0.0178} & 0.0466 & 0.0221 & 0.0323 & 0.0184 & 0.0183 \\
		T850 & 60h & 0.0251 & 0.0620 & 0.0299 & 0.0547 & 0.0255 & \textbf{0.0251} \\
		T850 & 72h & 0.0373 & 0.0835 & 0.0443 & 0.0716 & 0.0372 & \textbf{0.0356} \\
		T850 & 84h & 0.0573 & 0.1176 & 0.0607 & 0.1101 & 0.0561 & \textbf{0.0524} \\
		T850 & 96h & 0.0893 & 0.1698 & 0.0994 & 0.1556 & 0.0876 & \textbf{0.0795} \\
		T850 & 108h & 0.1375 & 0.2443 & 0.1423 & 0.2218 & 0.1364 & \textbf{0.1235} \\
		T850 & 120h & 0.2059 & 0.3395 & 0.2291 & 0.3035 & 0.2074 & \textbf{0.1919} \\
		T850 & 132h & 0.2959 & 0.4561 & 0.3103 & 0.4021 & \textbf{0.2714} & 0.2779 \\
		T850 & 144h & 0.4186 & 0.5987 & 0.4553 & 0.5464 & \textbf{0.3716} & 0.3909 \\
		T850 & 156h & 0.5656 & 0.7738 & 0.5670 & 0.7006 & \textbf{0.4904} & 0.5290 \\
		T850 & 168h & 0.7351 & 0.9761 & 0.7595 & 0.9217 & \textbf{0.6140} & 0.6940 \\
		T850 & 180h & 0.9111 & 1.1959 & 0.9136 & 1.1105 & \textbf{0.7442} & 0.8872 \\
		T850 & 192h & 1.1169 & 1.4065 & 1.1541 & 1.3722 & \textbf{0.8763} & 1.0970 \\
		T850 & 204h & 1.3242 & 1.6228 & 1.3345 & 1.5895 & \textbf{1.0046} & 1.3022 \\
		T850 & 216h & 1.5543 & 1.8264 & 1.5862 & 1.8438 & \textbf{1.1379} & 1.5319 \\
		T850 & 228h & 1.7904 & 2.0273 & 1.7694 & 2.0041 & \textbf{1.2647} & 1.7675 \\
		T850 & 240h & 2.0216 & 2.2219 & 2.0619 & 2.2228 & \textbf{1.3857} & 1.9841 \\
		\midrule
		Z500 & 12h & 6.1246 & 19.7576 & 6.5390 & 26.6958 & \textbf{5.4692} & 8.3412 \\
		Z500 & 24h & 32.5049 & 83.0658 & 35.6558 & 119.9938 & 27.7501 & \textbf{19.6303} \\
		Z500 & 36h & 143.1572 & 279.3724 & 140.8770 & 373.2934 & 120.7857 & \textbf{93.2486} \\
		Z500 & 48h & 422.6683 & 763.1224 & 392.4690 & 853.2862 & 369.6252 & \textbf{299.1150} \\
		Z500 & 60h & 987.4937 & 1701.4127 & 949.6804 & 1737.7194 & 876.4381 & \textbf{762.7267} \\
		Z500 & 72h & 1922.5214 & 3308.0522 & 1841.8360 & 3359.2755 & 1730.4532 & \textbf{1597.9239} \\
		Z500 & 84h & 3460.4605 & 5910.8440 & 3581.2083 & 5685.9188 & 3114.8778 & \textbf{3035.5692} \\
		Z500 & 96h & 5878.2943 & 9701.7932 & 5924.0644 & 9238.9938 & 5365.4816 & \textbf{5199.0114} \\
		Z500 & 108h & 9207.0815 & 14545.4886 & 9712.0473 & 14644.0367 & 8691.7610 & \textbf{8420.0934} \\
		Z500 & 120h & 13458.0770 & 20514.0618 & 14100.7981 & 20498.3937 & 13073.7371 & \textbf{12613.6073} \\
		Z500 & 132h & 19099.0530 & 28397.0361 & 20510.6848 & 27854.1528 & 18151.6719 & \textbf{17898.7947} \\
		Z500 & 144h & 26886.4435 & 38804.6110 & 27905.5223 & 38417.8133 & \textbf{24577.2698} & 24711.4568 \\
		Z500 & 156h & 36642.9245 & 51445.5371 & 38171.4196 & 51472.3563 & \textbf{32413.7343} & 33955.2442 \\
		Z500 & 168h & 48409.7383 & 65480.8310 & 48605.6695 & 66563.1440 & \textbf{41054.8529} & 45401.1447 \\
		Z500 & 180h & 60518.6516 & 80059.5577 & 62042.4154 & 82184.3559 & \textbf{49440.1657} & 58908.7073 \\
		Z500 & 192h & 72613.8939 & 93940.8663 & 74645.1235 & 94987.2438 & \textbf{56671.8077} & 74045.3921 \\
		Z500 & 204h & 85533.4150 & 108180.5635 & 89582.7398 & 107801.6592 & \textbf{63250.6773} & 90607.4881 \\
		Z500 & 216h & 100214.9498 & 122220.9953 & 103077.6203 & 118576.9672 & \textbf{69991.3504} & 106317.4791 \\
		Z500 & 228h & 116333.6458 & 137609.8284 & 120685.8377 & 131908.1464 & \textbf{76345.2022} & 122233.5907 \\
		Z500 & 240h & 131597.0478 & 153879.3978 & 135534.6276 & 143461.2929 & \textbf{82097.1713} & 137038.3218 \\
		\bottomrule
	\end{tabular}
\end{table}

\begin{table}[H]
	\caption{Variance of per-strata RMSE for global subregion. Lower is more fair; most fair for each variable and lead time is bolded. Smallest value determined before rounding to fourth decimal digit for display.}
	\label{subregion-variance_rmse_weighted_l2-benchmark}
	\scriptsize
	\centering
	\begin{tabular}{lllllllll}
		\\ \toprule
		& & \multicolumn{6}{c}{Model} \\
		\cmidrule(r){3-8}
		Variable & Lead time (h) & GraphCast & Keisler & Pangu-Weather & Spherical CNN & FuXi & NeuralGCM \\
		\midrule
		T850 & 12h & \textbf{0.0040} & 0.0191 & 0.0074 & 0.0108 & 0.0040 & 0.0054 \\
		T850 & 24h & \textbf{0.0058} & 0.0195 & 0.0083 & 0.0128 & 0.0058 & 0.0073 \\
		T850 & 36h & 0.0076 & 0.0266 & 0.0123 & 0.0170 & \textbf{0.0075} & 0.0093 \\
		T850 & 48h & 0.0099 & 0.0321 & 0.0142 & 0.0237 & \textbf{0.0096} & 0.0112 \\
		T850 & 60h & 0.0147 & 0.0432 & 0.0184 & 0.0326 & \textbf{0.0140} & 0.0150 \\
		T850 & 72h & 0.0240 & 0.0603 & 0.0297 & 0.0528 & 0.0226 & \textbf{0.0225} \\
		T850 & 84h & 0.0400 & 0.0885 & 0.0417 & 0.0735 & 0.0377 & \textbf{0.0355} \\
		T850 & 96h & 0.0654 & 0.1291 & 0.0742 & 0.1152 & 0.0627 & \textbf{0.0577} \\
		T850 & 108h & 0.1031 & 0.1828 & 0.1060 & 0.1500 & 0.0997 & \textbf{0.0917} \\
		T850 & 120h & 0.1575 & 0.2533 & 0.1753 & 0.2296 & 0.1554 & \textbf{0.1441} \\
		T850 & 132h & 0.2307 & 0.3424 & 0.2346 & 0.2908 & \textbf{0.2041} & 0.2146 \\
		T850 & 144h & 0.3239 & 0.4540 & 0.3420 & 0.4245 & \textbf{0.2741} & 0.3044 \\
		T850 & 156h & 0.4354 & 0.5892 & 0.4239 & 0.5038 & \textbf{0.3547} & 0.4125 \\
		T850 & 168h & 0.5633 & 0.7316 & 0.5709 & 0.6931 & \textbf{0.4451} & 0.5456 \\
		T850 & 180h & 0.7000 & 0.8827 & 0.6815 & 0.8021 & \textbf{0.5407} & 0.6897 \\
		T850 & 192h & 0.8463 & 1.0458 & 0.8625 & 1.0085 & \textbf{0.6392} & 0.8390 \\
		T850 & 204h & 0.9986 & 1.2096 & 0.9686 & 1.0821 & \textbf{0.7371} & 0.9964 \\
		T850 & 216h & 1.1588 & 1.3546 & 1.1507 & 1.2710 & \textbf{0.8334} & 1.1658 \\
		T850 & 228h & 1.3176 & 1.4905 & 1.2595 & 1.3367 & \textbf{0.9237} & 1.3212 \\
		T850 & 240h & 1.4755 & 1.6207 & 1.4524 & 1.5565 & \textbf{1.0082} & 1.4636 \\
		\midrule
		Z500 & 12h & 4.3923 & 21.0451 & \textbf{2.8972} & 13.7473 & 4.1863 & 7.5274 \\
		Z500 & 24h & 20.3463 & 51.3661 & 23.1773 & 69.7545 & 16.2691 & \textbf{11.0202} \\
		Z500 & 36h & 100.9786 & 180.1252 & 103.6128 & 257.4925 & 81.9689 & \textbf{58.8206} \\
		Z500 & 48h & 306.4674 & 508.7010 & 311.7182 & 654.5374 & 260.7049 & \textbf{211.2787} \\
		Z500 & 60h & 716.7357 & 1184.5767 & 745.8135 & 1328.9283 & 627.4897 & \textbf{550.1721} \\
		Z500 & 72h & 1409.4286 & 2363.1666 & 1476.1021 & 2530.5781 & 1266.1543 & \textbf{1185.5958} \\
		Z500 & 84h & 2530.5646 & 4270.6684 & 2772.6681 & 4229.7121 & 2300.8085 & \textbf{2260.5735} \\
		Z500 & 96h & 4226.6600 & 7043.0612 & 4561.6586 & 6801.1818 & 3926.2459 & \textbf{3892.6980} \\
		Z500 & 108h & 6720.7189 & 10643.8530 & 7468.2495 & 10438.1276 & 6378.4510 & \textbf{6359.9607} \\
		Z500 & 120h & 10141.6595 & 15290.3272 & 10987.4233 & 15390.8119 & \textbf{9782.2574} & 9793.6826 \\
		Z500 & 132h & 14620.2634 & 21372.4804 & 15893.7948 & 21336.5717 & \textbf{13540.9848} & 14039.6015 \\
		Z500 & 144h & 20488.9905 & 29376.4397 & 21538.2609 & 28923.9358 & \textbf{18291.9364} & 19539.8322 \\
		Z500 & 156h & 27792.6869 & 39121.5091 & 29073.4592 & 38107.4375 & \textbf{24161.9448} & 26805.0344 \\
		Z500 & 168h & 36070.0516 & 49415.4562 & 37029.6547 & 49476.3792 & \textbf{30594.7232} & 35135.2208 \\
		Z500 & 180h & 44563.2319 & 59680.5409 & 46715.6210 & 59530.9165 & \textbf{36729.6627} & 44896.8645 \\
		Z500 & 192h & 53264.2736 & 69510.8777 & 55549.5099 & 69291.2225 & \textbf{42029.8539} & 56232.0981 \\
		Z500 & 204h & 62941.2933 & 80014.6105 & 65986.8602 & 78463.7736 & \textbf{47054.4341} & 69073.4405 \\
		Z500 & 216h & 73576.4355 & 90367.1381 & 76461.9376 & 88426.2503 & \textbf{52021.9554} & 81403.3079 \\
		Z500 & 228h & 84684.8208 & 101170.3770 & 88451.4026 & 99330.6919 & \textbf{56592.5334} & 92347.6782 \\
		Z500 & 240h & 95626.2297 & 111864.2591 & 98694.2737 & 109405.8148 & \textbf{60587.5795} & 102322.9160 \\
		\bottomrule
	\end{tabular}
\end{table}

\begin{table}[H]
	\caption{Variance of per-strata RMSE for income. Lower is more fair; most fair for each variable and lead time is bolded. Smallest value determined before rounding to fourth decimal digit for display.}
	\label{income-variance_rmse_weighted_l2-benchmark}
	\scriptsize
	\centering
	\begin{tabular}{lllllllll}
		\\ \toprule
		& & \multicolumn{6}{c}{Model} \\
		\cmidrule(r){3-8}
		Variable & Lead time (h) & GraphCast & Keisler & Pangu-Weather & Spherical CNN & FuXi & NeuralGCM \\
		\midrule
		T850 & 12h & 0.0006 & 0.0013 & 0.0009 & 0.0011 & 0.0007 & \textbf{0.0006} \\
		T850 & 24h & 0.0010 & 0.0016 & 0.0011 & 0.0017 & 0.0011 & \textbf{0.0009} \\
		T850 & 36h & 0.0011 & 0.0019 & 0.0013 & 0.0016 & 0.0012 & \textbf{0.0010} \\
		T850 & 48h & 0.0013 & 0.0033 & 0.0016 & 0.0056 & 0.0014 & \textbf{0.0010} \\
		T850 & 60h & 0.0026 & 0.0061 & 0.0026 & 0.0057 & 0.0025 & \textbf{0.0018} \\
		T850 & 72h & 0.0062 & 0.0125 & 0.0073 & 0.0212 & 0.0059 & \textbf{0.0044} \\
		T850 & 84h & 0.0139 & 0.0243 & 0.0145 & 0.0273 & 0.0135 & \textbf{0.0106} \\
		T850 & 96h & 0.0276 & 0.0438 & 0.0312 & 0.0603 & 0.0265 & \textbf{0.0227} \\
		T850 & 108h & 0.0488 & 0.0724 & 0.0543 & 0.0721 & 0.0478 & \textbf{0.0426} \\
		T850 & 120h & 0.0798 & 0.1103 & 0.0893 & 0.1264 & 0.0785 & \textbf{0.0716} \\
		T850 & 132h & 0.1238 & 0.1542 & 0.1313 & 0.1532 & \textbf{0.1045} & 0.1113 \\
		T850 & 144h & 0.1762 & 0.2094 & 0.1835 & 0.2357 & \textbf{0.1380} & 0.1600 \\
		T850 & 156h & 0.2393 & 0.2702 & 0.2485 & 0.2810 & \textbf{0.1782} & 0.2193 \\
		T850 & 168h & 0.3072 & 0.3337 & 0.3235 & 0.3675 & \textbf{0.2187} & 0.2877 \\
		T850 & 180h & 0.3717 & 0.3958 & 0.3883 & 0.4064 & \textbf{0.2567} & 0.3626 \\
		T850 & 192h & 0.4403 & 0.4646 & 0.4587 & 0.4712 & \textbf{0.2931} & 0.4383 \\
		T850 & 204h & 0.5147 & 0.5308 & 0.5126 & 0.5191 & \textbf{0.3260} & 0.5095 \\
		T850 & 216h & 0.5801 & 0.5981 & 0.5677 & 0.5882 & \textbf{0.3587} & 0.5835 \\
		T850 & 228h & 0.6361 & 0.6526 & 0.6199 & 0.6570 & \textbf{0.3897} & 0.6473 \\
		T850 & 240h & 0.6864 & 0.6947 & 0.6519 & 0.7183 & \textbf{0.4179} & 0.6952 \\
		\midrule
		Z500 & 12h & \textbf{0.1028} & 2.0068 & 0.4295 & 4.7712 & 0.3603 & 0.6834 \\
		Z500 & 24h & 8.1396 & 11.1313 & 11.5812 & 32.1529 & 9.3156 & \textbf{3.6432} \\
		Z500 & 36h & 58.2438 & 50.2198 & 68.4765 & 189.2996 & 56.8211 & \textbf{36.4330} \\
		Z500 & 48h & 189.4209 & 184.8858 & 223.4966 & 442.7660 & 179.4246 & \textbf{136.8048} \\
		Z500 & 60h & 451.0046 & 520.4977 & 519.4968 & 974.0071 & 428.2830 & \textbf{358.7665} \\
		Z500 & 72h & 911.0966 & 1229.6501 & 1080.3229 & 1821.1763 & 883.6421 & \textbf{785.7836} \\
		Z500 & 84h & 1646.9627 & 2382.9868 & 1968.3536 & 2952.4151 & 1629.0901 & \textbf{1513.8882} \\
		Z500 & 96h & 2737.5665 & 4144.1104 & 3276.2421 & 4528.4890 & 2747.9434 & \textbf{2587.6919} \\
		Z500 & 108h & 4348.0463 & 6467.1286 & 5135.8483 & 6914.8034 & 4382.9351 & \textbf{4201.6065} \\
		Z500 & 120h & 6597.2991 & 9373.6499 & 7560.5792 & 9862.9344 & 6573.8777 & \textbf{6412.9097} \\
		Z500 & 132h & 9533.7548 & 12964.9450 & 10791.6058 & 13911.4043 & \textbf{8959.9675} & 9339.7532 \\
		Z500 & 144h & 13148.0557 & 17427.2140 & 14753.4857 & 18254.7028 & \textbf{11740.4081} & 12892.9825 \\
		Z500 & 156h & 17445.2287 & 22671.1704 & 19650.6572 & 22708.4878 & \textbf{14994.6352} & 17418.3039 \\
		Z500 & 168h & 22182.4991 & 28490.4680 & 25084.7184 & 28230.3578 & \textbf{18601.2094} & 22431.9354 \\
		Z500 & 180h & 27146.0603 & 34351.0365 & 31108.3918 & 34170.3147 & \textbf{22273.9916} & 28154.0990 \\
		Z500 & 192h & 32239.4210 & 40199.3865 & 36779.7222 & 38994.5403 & \textbf{25839.6953} & 34533.8555 \\
		Z500 & 204h & 37995.0056 & 46201.5532 & 42988.3711 & 45546.9357 & \textbf{29384.0512} & 41433.0906 \\
		Z500 & 216h & 43861.1521 & 52094.4753 & 48642.4069 & 51911.1326 & \textbf{32665.9497} & 48114.7002 \\
		Z500 & 228h & 49661.7138 & 57430.4755 & 54233.6775 & 58344.7264 & \textbf{35546.9475} & 54187.6957 \\
		Z500 & 240h & 54749.8350 & 62421.6302 & 58373.7147 & 63053.7066 & \textbf{38117.6736} & 60038.5224 \\
		\bottomrule
	\end{tabular}
\end{table}

\begin{table}[H]
	\caption{Variance of per-strata RMSE for landcover. Lower is more fair; most fair for each variable and lead time is bolded. Smallest value determined before rounding to fourth decimal digit for display.}
	\label{landcover-variance_rmse_weighted_l2-benchmark}
	\scriptsize
	\centering
	\begin{tabular}{lllllllll}
		\\ \toprule
		& & \multicolumn{6}{c}{Model} \\
		\cmidrule(r){3-8}
		Variable & Lead time (h) & GraphCast & Keisler & Pangu-Weather & Spherical CNN & FuXi & NeuralGCM \\
		\midrule
		T850 & 12h & 0.0000 & 0.0000 & 0.0000 & \textbf{0.0000} & 0.0002 & 0.0002 \\
		T850 & 24h & 0.0007 & 0.0002 & 0.0007 & 0.0003 & 0.0011 & \textbf{0.0001} \\
		T850 & 36h & 0.0014 & \textbf{0.0002} & 0.0012 & 0.0005 & 0.0018 & 0.0006 \\
		T850 & 48h & 0.0020 & \textbf{0.0004} & 0.0017 & 0.0010 & 0.0026 & 0.0014 \\
		T850 & 60h & 0.0024 & \textbf{0.0005} & 0.0027 & 0.0015 & 0.0032 & 0.0021 \\
		T850 & 72h & 0.0028 & \textbf{0.0007} & 0.0029 & 0.0022 & 0.0037 & 0.0029 \\
		T850 & 84h & 0.0030 & \textbf{0.0008} & 0.0042 & 0.0027 & 0.0040 & 0.0033 \\
		T850 & 96h & 0.0030 & \textbf{0.0008} & 0.0036 & 0.0030 & 0.0043 & 0.0036 \\
		T850 & 108h & 0.0027 & \textbf{0.0006} & 0.0043 & 0.0030 & 0.0040 & 0.0034 \\
		T850 & 120h & 0.0021 & \textbf{0.0004} & 0.0032 & 0.0029 & 0.0035 & 0.0028 \\
		T850 & 132h & 0.0013 & \textbf{0.0002} & 0.0033 & 0.0027 & 0.0018 & 0.0019 \\
		T850 & 144h & 0.0006 & \textbf{0.0000} & 0.0019 & 0.0017 & 0.0010 & 0.0012 \\
		T850 & 156h & 0.0002 & \textbf{0.0000} & 0.0016 & 0.0014 & 0.0004 & 0.0007 \\
		T850 & 168h & \textbf{0.0000} & 0.0001 & 0.0005 & 0.0008 & 0.0001 & 0.0004 \\
		T850 & 180h & 0.0002 & 0.0003 & 0.0006 & 0.0005 & \textbf{0.0000} & 0.0001 \\
		T850 & 192h & 0.0005 & 0.0008 & 0.0001 & 0.0001 & 0.0000 & \textbf{0.0000} \\
		T850 & 204h & 0.0011 & 0.0017 & 0.0004 & \textbf{0.0001} & 0.0001 & 0.0001 \\
		T850 & 216h & 0.0017 & 0.0028 & 0.0001 & \textbf{0.0000} & 0.0002 & 0.0003 \\
		T850 & 228h & 0.0025 & 0.0036 & 0.0002 & \textbf{0.0001} & 0.0003 & 0.0006 \\
		T850 & 240h & 0.0039 & 0.0043 & \textbf{0.0000} & 0.0006 & 0.0004 & 0.0011 \\
		\midrule
		Z500 & 12h & 0.3305 & 4.0325 & \textbf{0.2388} & 1.5367 & 0.6630 & 0.8439 \\
		Z500 & 24h & \textbf{1.5799} & 8.5578 & 1.8622 & 6.6101 & 2.8439 & 2.2000 \\
		Z500 & 36h & \textbf{6.6135} & 20.1753 & 8.5065 & 23.4304 & 9.5159 & 8.2295 \\
		Z500 & 48h & \textbf{22.2078} & 47.9855 & 25.5133 & 54.7201 & 27.5102 & 22.8971 \\
		Z500 & 60h & 55.5015 & 107.3930 & 65.5437 & 118.5960 & 64.4349 & \textbf{52.2397} \\
		Z500 & 72h & 114.5290 & 200.5459 & 125.8643 & 203.6786 & 129.0179 & \textbf{101.3914} \\
		Z500 & 84h & 208.7222 & 324.2265 & 224.6944 & 355.2449 & 228.5063 & \textbf{184.4539} \\
		Z500 & 96h & 346.5953 & 490.1982 & 348.0649 & 542.9503 & 371.9671 & \textbf{296.5129} \\
		Z500 & 108h & 516.4417 & 709.8745 & 526.3853 & 830.3223 & 558.4856 & \textbf{440.8835} \\
		Z500 & 120h & 707.6771 & 998.0250 & 777.2581 & 1187.9308 & 792.2084 & \textbf{622.6550} \\
		Z500 & 132h & 940.2671 & 1344.0848 & 1099.8318 & 1611.4985 & 1010.1114 & \textbf{866.4512} \\
		Z500 & 144h & 1212.0642 & 1702.3716 & 1454.6969 & 2119.7129 & 1301.3511 & \textbf{1163.0654} \\
		Z500 & 156h & 1495.3124 & 2073.6823 & 1866.4296 & 2735.7801 & 1609.5192 & \textbf{1494.8609} \\
		Z500 & 168h & \textbf{1848.1638} & 2508.5428 & 2369.1183 & 3325.6011 & 1895.0281 & 1955.8816 \\
		Z500 & 180h & 2325.0811 & 3001.6319 & 3056.2699 & 4049.4052 & \textbf{2237.4910} & 2512.1880 \\
		Z500 & 192h & 2904.0572 & 3531.7694 & 3808.4100 & 4735.2684 & \textbf{2719.2835} & 3055.4490 \\
		Z500 & 204h & 3500.5399 & 4080.5920 & 4721.5363 & 5753.4311 & \textbf{3302.9746} & 3664.9805 \\
		Z500 & 216h & 4025.9693 & 4742.4189 & 5539.3190 & 6689.3969 & \textbf{3904.1884} & 4354.9488 \\
		Z500 & 228h & \textbf{4442.7599} & 5335.4729 & 6527.9828 & 7811.2172 & 4488.5863 & 5085.2891 \\
		Z500 & 240h & \textbf{4883.8882} & 5846.0681 & 7480.0098 & 8736.5929 & 5066.5870 & 5873.1976 \\
		\bottomrule
	\end{tabular}
\end{table}

\section{Supplemental figures}

\begin{figure}
  \centering
  \includegraphics[width=0.7\linewidth]{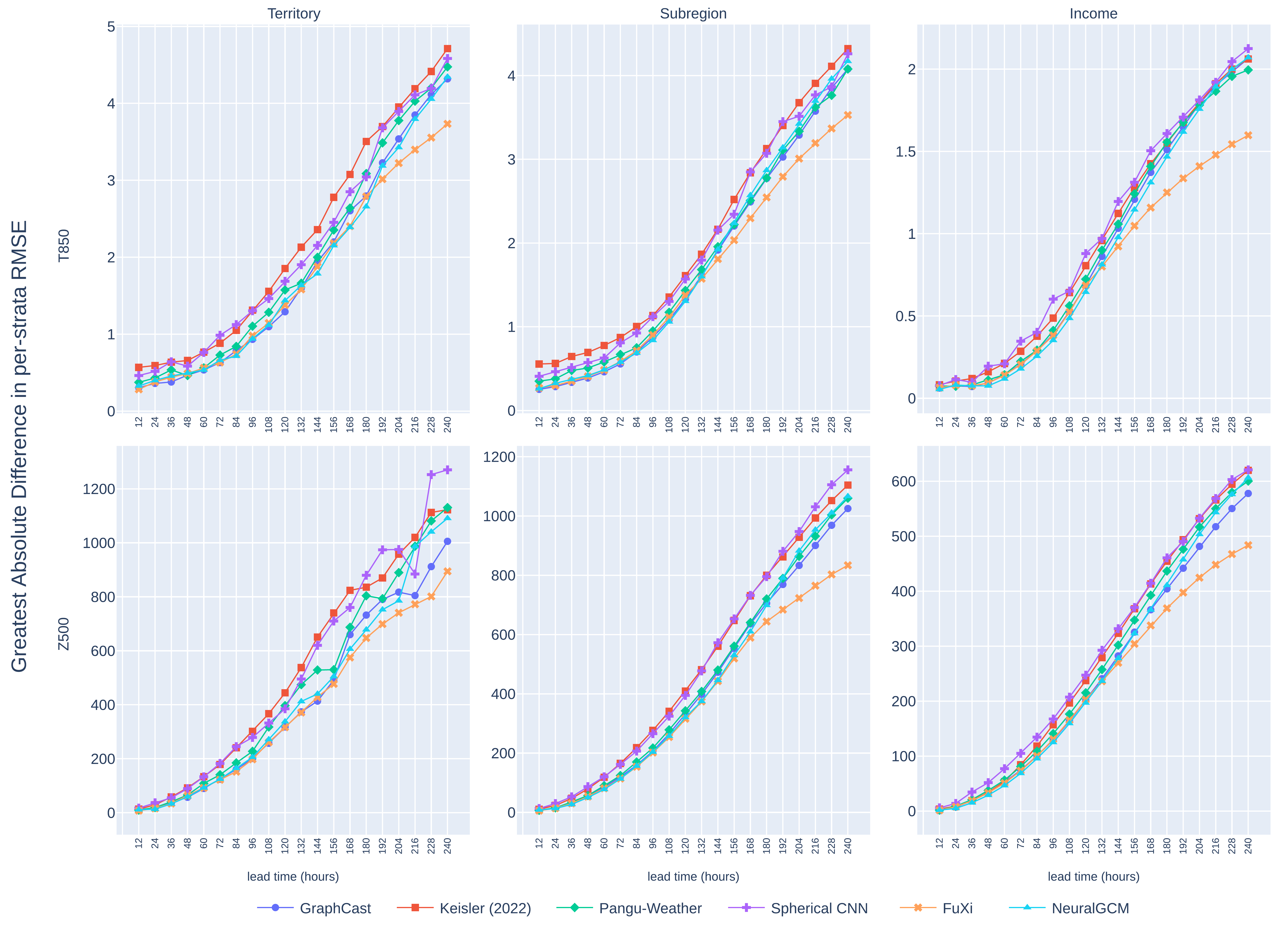}
  \caption{Greatest absolute difference of any two per-strata RMSE for each attribute when predicting T850 and Z500 at different lead times. Lower difference is more fair. Outlier RMSE values have been removed. Starting at a lead time of one week, FuXi is still the most fair model across all attributes and variables.}
  \label{rmse-diff-no-outliers}
\end{figure}

\begin{figure}
  \centering
  \includegraphics[width=0.7\linewidth]{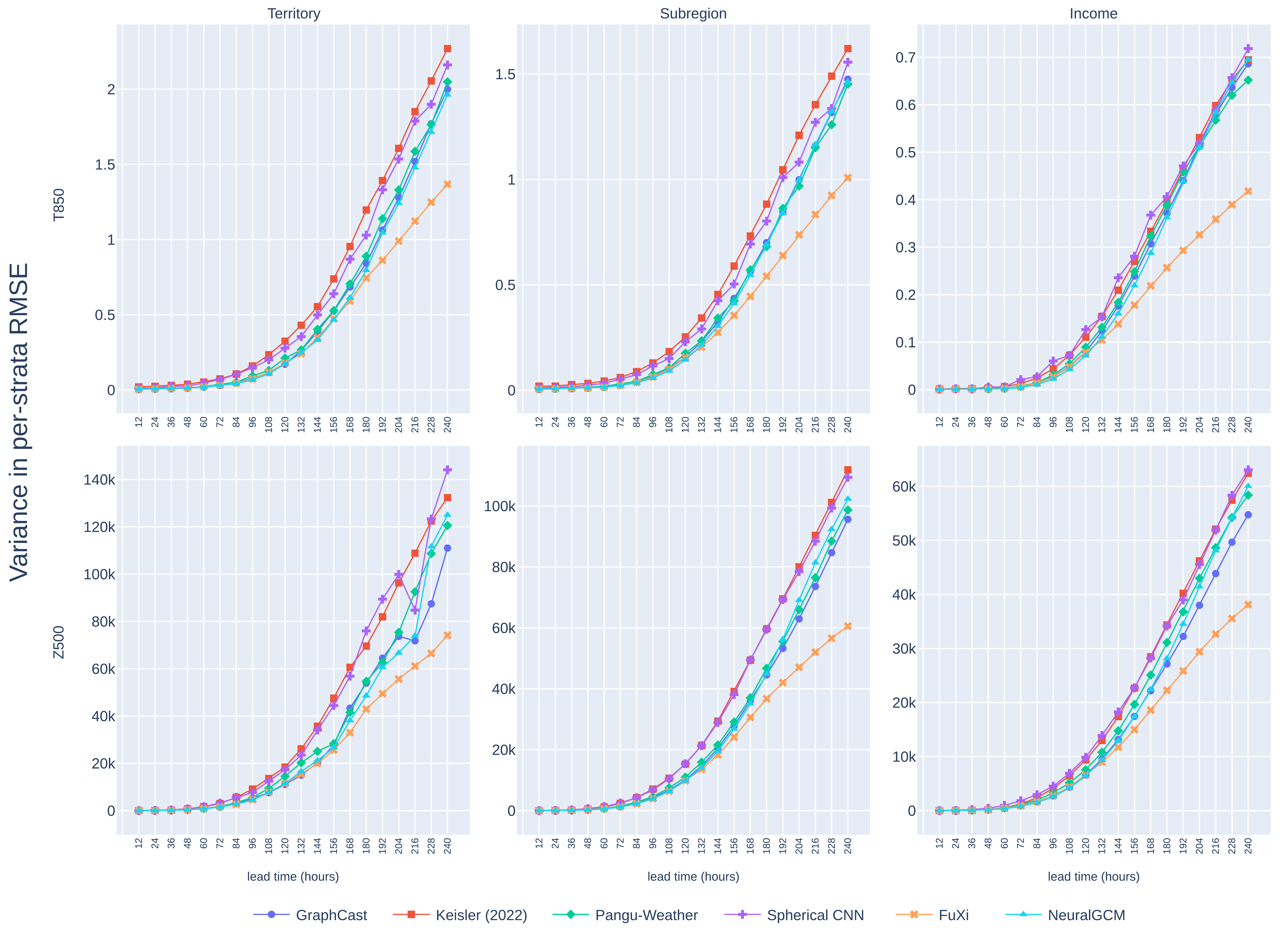}
  \caption{Variance of all the per-strata RMSE for each attribute when predicting T850 and Z500 at different lead times. Lower variance is more fair. Outlier RMSE values have been removed. Starting at a lead time of one week, FuXi is still the most fair model across all attributes and variables.}
  \label{rmse-var-no-outliers}
\end{figure}

\begin{figure}
  \centering
  \includegraphics[width=0.7\linewidth]{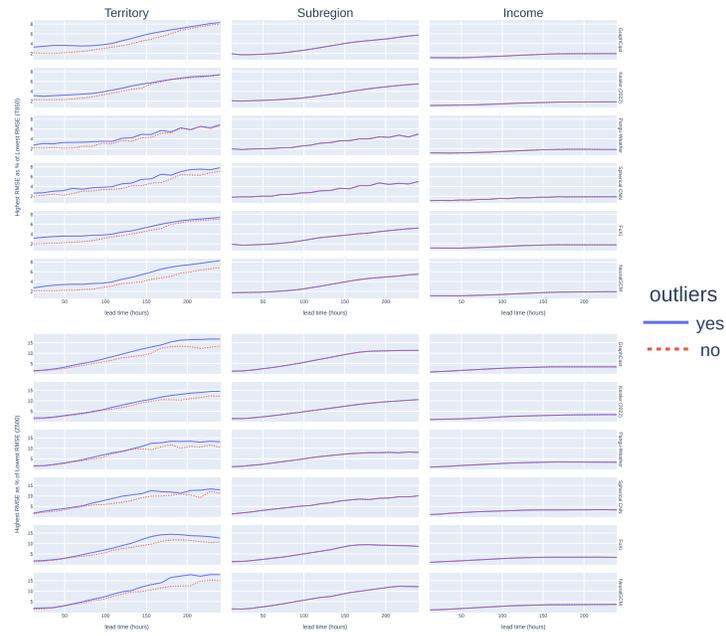}
  \caption{Highest per-strata RMSE as a percent of the lowest per-strata RMSE with and without RMSE outliers filtered out.}
  \label{rmse-percent}
\end{figure}

\begin{figure}[!htb]
  \centering
  \includegraphics[width=0.7\linewidth]{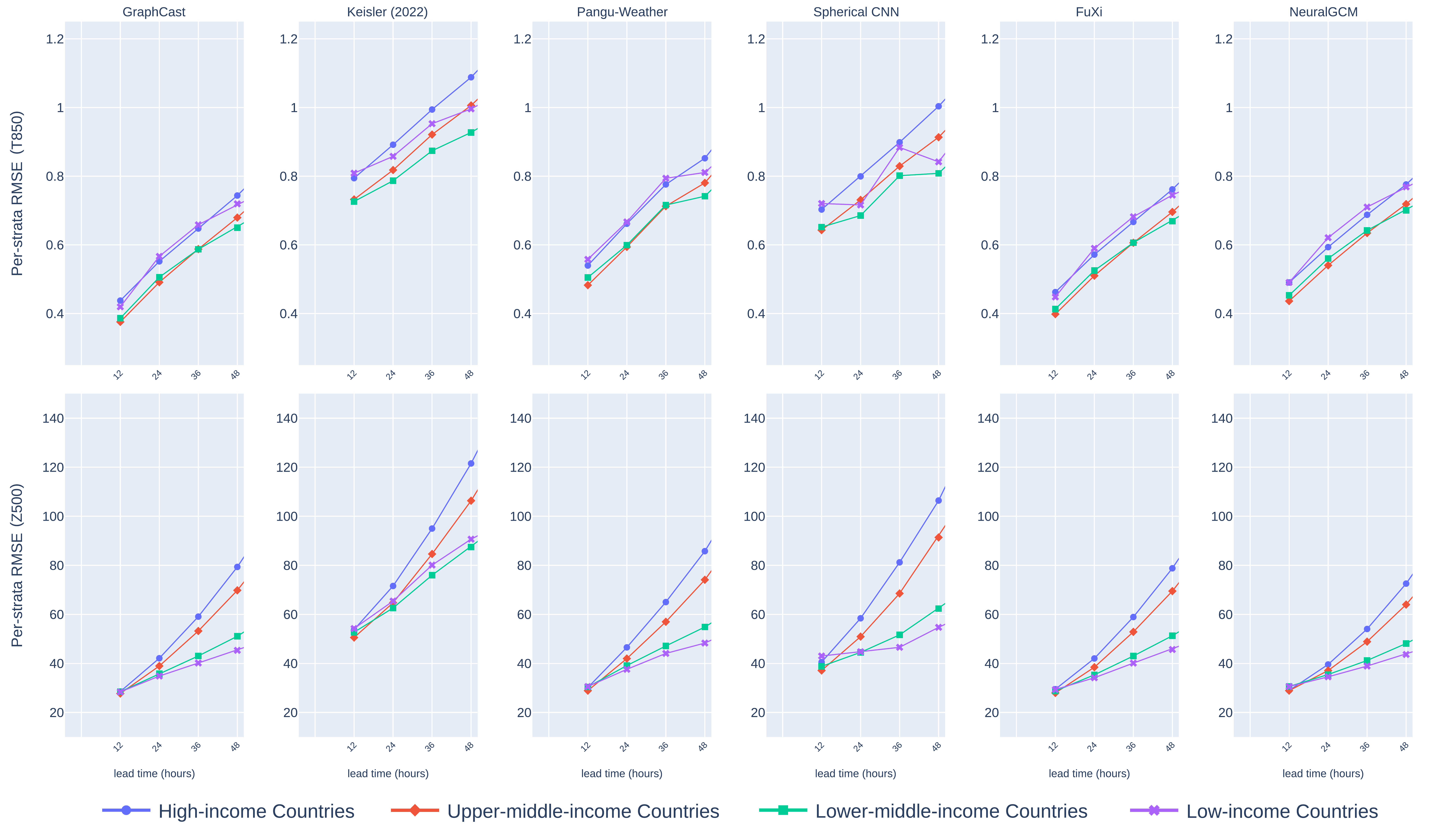}
  \caption{Per-strata RMSE for the income attribute of each model for the first 48 hours of lead time.}
  \label{fig-zoomed-income}
\end{figure}

\end{document}